%% file: main.tex
\documentclass[10pt,twocolumn,letterpaper]{article}
\usepackage[pagenumbers]{paperstyle}
\input{preamble}
\definecolor{linkblue}{rgb}{0.21,0.49,0.74}
\usepackage[breaklinks,colorlinks,allcolors=linkblue]{hyperref}

\def\paperID{}
\def\confName{arXiv}
\def\confYear{2026}
\newcommand{\suppref}{the appendix}
\makeatletter  %
\patchcmd{\@maketitle}{\vskip .375in}{\vskip .05in}{}{\PackageWarning{main}{title patch 1 failed}}
\patchcmd{\@maketitle}{{\vspace*{24pt}}}{{\vspace*{14pt}}}{}{\PackageWarning{main}{title patch 2 failed}}
\patchcmd{\@maketitle}{\vspace*{12pt}}{\vspace*{2pt}}{}{\PackageWarning{main}{title patch 3 failed}}
\makeatother

\title{How Much Can Best-of-N Help a World Action Model?\\Measuring Selectable Headroom and Realizing It in Closed Loop}
\author{
Haojia Gao\textsuperscript{1,2}\quad
Yue Cao\textsuperscript{2,3,4}\quad
Yumin Liu\textsuperscript{1}\quad
Haohua Que\textsuperscript{1}\quad
Mingkai Liu\textsuperscript{5}\\
Guanqiu Guo\textsuperscript{1,2,4}\quad
Zhuo Zou\textsuperscript{3}\quad
Bokui Chen\textsuperscript{1}\quad
Xueqian Wang\textsuperscript{1,2,4}\thanks{Corresponding author: \texttt{wang.xq@sz.tsinghua.edu.cn}}\\[4pt]
{\normalsize\textsuperscript{1}Tsinghua University\quad
\textsuperscript{2}Everwise-Tech Co., Ltd.\quad
\textsuperscript{3}Fudan University}\\
{\normalsize\textsuperscript{4}Research Institute of Tsinghua University in Shenzhen\quad
\textsuperscript{5}Peking University}
}

\begin{document}
\maketitle
\input{body}
\clearpage  %
{
    \small
    \bibliography{refs}
}

\clearpage
\input{appendix}
\end{document}

%% file: preamble.tex
\usepackage{bm}
\usepackage{multirow}
\usepackage{tikz}
\usetikzlibrary{arrows.meta,positioning}
\newcommand{\Oone}{\bar O^{1}}
\newcommand{\Oin}{\bar O^{\mathrm{in}}}
\newcommand{\OL}{\bar O^{L}}
\newcommand{\phat}{\hat p}
\newcommand{\tauhat}{\hat\tau^{2}}

%% file: body.tex
\begin{abstract}
World action models (WAMs) predict the visual consequences of the actions they generate, which makes test-time selection a natural way to improve a fixed policy: imagine the futures of candidate actions and execute the best one. The benefit of selection is bounded by the \emph{selectable headroom}, how much better the best candidate truly is, but the standard estimate, which executes each candidate once and keeps the best outcome, conflates candidate quality with the randomness of everything that follows. We introduce a denoised counterfactual protocol, based on bitwise-reproducible branching, repeated continuations, held-out oracles, and common random numbers, that measures the headroom a selector can realize, and use its measurements to build a fail-gated selector. Across four WAMs, two benchmarks, over 100{,}000 branched rollouts, and pre-registered tests, we find that (i) the standard estimate overstates the selectable headroom of native samples: bounds from 17 continuations per candidate place its selectable share between 24\% and 66\% for three WAMs; (ii) structured action perturbations create up to 16 points of selectable headroom; and (iii) in both WAMs with action-conditioned imagination, imagined futures carry candidate-outcome information with no detectable loss relative to real futures, information that the value head of Cosmos Policy leaves unused. Deviating from the policy chunk only when a readout of the WAM predicts failure, the selector raises closed-loop task success from 76.9\% to 81.6\% on held-out episodes of training tasks, with the largest gain where the policy fails most often, and from 62.6\% to 69.8\% on unseen tasks.
\end{abstract}

\section{Introduction}\label{sec:intro}

World action models, which extend video-prediction policies and joint video-action models~\citep{du2023unipi,wu2024gr1,hu2025vpp,li2025uva,zhu2025uwm}, jointly produce an action chunk, a prediction of the scene after the chunk is executed, and an estimate of its value~\citep{kim2026cosmospolicy}. This structure makes a simple form of test-time compute available to any WAM: sample several candidate chunks, imagine where each of them leads, and execute the chunk whose imagined future looks best. Sampling-and-verification schemes of this kind, established for reasoning with language models~\citep{cobbe2021verifiers,lightman2024verify,snell2025scaling}, have improved vision-language-action (VLA~\citep{kim2024openvla,black2024pi0}) policies~\citep{kwok2025robomonkey,nakamoto2024vgps,jang2025verifierfree,kwok2026scaling}, and verifiers specific to WAMs have been proposed~\citep{zhao2026geobon,zhang2026wcd}.

How much can such selection gain? The answer is fixed before any scorer is chosen: the gain is bounded by how much better the best candidate in the pool truly is. We call this quantity the \emph{selectable headroom}. \citet{yuan2026beyond} measure it directly with counterfactual branching. They restore the same simulator state, execute each of 16 sampled candidates, continue each candidate with the policy until the episode ends, and find that choosing the candidate with the best outcome would raise success from 68.9\% to 79.2\%, whereas learned values, image similarity, and VLM judgments recover little of this gain.

This estimate executes each candidate \emph{once}. Everything that happens after the candidate, including every later policy query, is random, so the candidate with the best realized outcome is partly the candidate whose continuation happened to go well. Selecting and scoring on the same random draw is the optimizer's curse~\citep{smith2006optimizer}, the same bias that double estimators remove in reinforcement learning~\citep{vanhasselt2010double}. The question that matters for selection is therefore how much of the oracle gain a selector could ever realize.

\begin{figure*}[t]
\centering
\includegraphics[width=\linewidth]{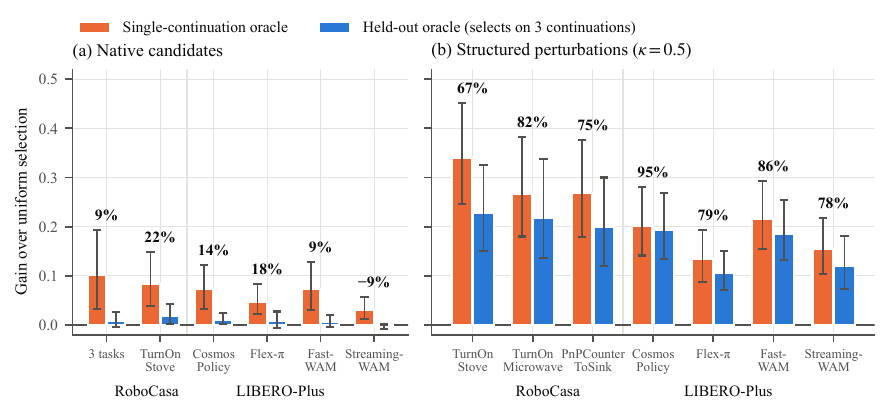}
\caption{\textbf{How much of the measured headroom is selectable?} Orange: the single-continuation oracle (best realized outcome). Blue: the held-out oracle, which selects on three continuations and is scored on a fresh one. Percentages: the held-out share $\OL/\Oone$, a lower bound on the selectable share. For native samples (a), this share is at most 22\% on both benchmarks and for all four models; for structured action perturbations (b), it is 67--95\%. Error bars: 95\% CIs from a bootstrap clustered by state.}
\label{fig:teaser}
\end{figure*}

We answer this question with a \textbf{denoised counterfactual protocol} (\cref{fig:overview}a). Every candidate is executed from a bitwise-reproducible simulator snapshot and continued several times. The realizable part of the headroom is measured by a held-out oracle that selects on some continuations and is scored on others; a variance decomposition separates the effect of the candidate from its interaction with the continuation; and every evaluator is run with common random numbers, so that all candidates are compared under the same evaluation randomness. Using more than 100{,}000 branched rollouts of four WAMs, namely Cosmos Policy~\citep{kim2026cosmospolicy}, Flex-$\pi$~\citep{yan2026flexpi}, Fast-WAM~\citep{yuan2026fastwam}, and Streaming-WAM~\citep{huang2026streamingwam}, on RoboCasa~\citep{nasiriany2024robocasa} and LIBERO-Plus~\citep{fei2025liberoplus}, we establish the following results, each supported by pre-registered tests.

\begin{itemize}
  \item \textbf{The single-continuation oracle is dominated by continuation randomness} (\cref{sec:noise}). Under the protocol of \citet{yuan2026beyond}, we reproduce a 12.3-point oracle gain, yet a selector that observes three independent outcomes per candidate realizes only 9\% of it on a fresh continuation, and 98\% of the candidate-specific outcome variation is re-drawn when the continuation seeds change. The same holds for all four WAMs on LIBERO-Plus. With 17 continuations per candidate, the selectable share lies between 24\% and 52\% of the single-continuation oracle for Cosmos Policy and between a third and two thirds for Flex-$\pi$ and Fast-WAM (\cref{fig:teaser}).
  \item \textbf{Structured proposals create selectable headroom} (\cref{sec:proposals}). Native samples are nearly identical and offer almost no selectable headroom over the policy chunk, whereas displacing the policy chunk along the robot axes creates headroom of which 67--95\% survives held-out evaluation for all four WAMs. This headroom lies where the policy chunk fails, and it exceeds the chunk by 4.5--16 points on both benchmarks and for all four WAMs.
  \item \textbf{The imagination carries the signal that the value head misses} (\cref{sec:imagination}). On both WAMs with action-conditioned imagination, a linear readout of the \emph{imagined} future ranks candidates with no detectable loss relative to the same readout of the \emph{real} future; for Cosmos Policy, the signal resides in the predicted consequence rather than in the encoding of the action. The value head, although accurate at the state level (AUROC 0.90), recovers almost none of this signal.
  \item \textbf{A selector built on the findings improves on the policy} (\cref{sec:selector}). A learned failure gate decides whether to deviate from the policy chunk, and a learned ranker decides where to go, so the selector deviates only where the policy is likely to fail, which is where the headroom lies, and executes the policy chunk unchanged everywhere else. Frozen and pre-registered, it raises closed-loop task success from 76.9\% to 81.6\% on reserved episodes of the training tasks, by 10.7 points where the policy fails most often, and, when re-trained without the test tasks, from 62.6\% to 69.8\% on unseen tasks.
\end{itemize}

These results show when test-time selection can help a WAM: the headroom worth pursuing is created by proposals that displace the policy chunk rather than by native samples, the imagined futures of the model already carry information that ranks these proposals, and what is missing is a readout trained to compare sibling actions rather than to value states. Together, the results give a concrete recipe for test-time compute in a WAM: spend it on consequential proposals at states where the policy is likely to fail, score them with readouts trained on branch outcomes, and use the held-out oracle, at the cost of one extra continuation per candidate, to know in advance how much a selector can gain. We package the protocol and the branching data as \textbf{HeadroomBench} (\cref{sec:bench}), which will be publicly released.

\section{Related work}\label{sec:related}

\paragraph{Test-time selection for robot policies.} RoboMonkey samples and perturbs VLA actions and selects among them with a VLM verifier~\citep{kwok2025robomonkey}; V-GPS re-ranks samples of generalist policies with an offline-RL value function~\citep{nakamoto2024vgps}; verifier-free sampling~\citep{jang2025verifierfree} and scaled verification~\citep{kwok2026scaling} offer further variants; and BOKBO adds calibrated abstention when all candidates are poor~\citep{singh2026bokbo}. For WAMs, Cosmos Policy plans with a fine-tuned value model on a real bimanual robot~\citep{kim2026cosmospolicy}; Motus2 exposes a policy, an action-conditioned simulator, and a value model within a single network~\citep{bi2026motus2}; GeoBoN scores imagined futures by multi-view geometric consistency~\citep{zhao2026geobon}; and World-Coherent Decoding ranks candidates by internal generative signals that are calibrated online against the mismatch between imagined and observed futures~\citep{zhang2026wcd}. KeyStone selects without a judge by returning the medoid of the dominant cluster of sampled chunks~\citep{dai2026keystone}. Other work uses the imagination to refine a proposed chunk ($\omega$-EVA~\citep{sun2026omegaeva}) or to decide when to stop executing it by comparing predicted and observed futures (adaptive execution~\citep{wang2026trust}). In language models, the coverage of repeated sampling keeps growing while selection without an automatic verifier plateaus~\citep{brown2024monkeys}, and best-of-$n$ selection over-optimizes a learned reward~\citep{gao2023overoptimization}. These works design selectors and verifiers; we measure the headroom for which they compete, identify where it comes from, and use these measurements to design a selector that improves the policy in closed loop.

\paragraph{Evaluating and diagnosing world models.} Learned world models have long served planning and control~\citep{finn2017foresight,ebert2018foresight,hafner2019planet,hansen2024tdmpc2,zhou2025dinowm}. Recent work evaluates the dynamic consistency of WAM predictions~\citep{ruan2026future,mehta2026imply}, guards policy optimization against model hallucination~\citep{chen2026hawmpo,li2026dreamledger}, argues for decision-centric evaluation of world models~\citep{yu2026position}, shows that imperfect world models are exploitable~\citep{bhamidipaty2026exploitable}, and evaluates real policies with world models or simulation~\citep{li2025worldeval,li2024simpler}. Closest to our work, \citet{yuan2026beyond} introduce same-state counterfactual branching for test-time planning with WAMs on the same model and simulator. We build on their design, adopt their protocol for a direct comparison, and add repeated continuations, held-out oracles, controlled proposals, and representation-level readouts. Our held-out oracle addresses the optimizer's curse~\citep{smith2006optimizer} in the spirit of double estimators~\citep{vanhasselt2010double}, our evaluators use common random numbers~\citep{glasserman1992crn,ng2000pegasus}, and our key tests were pre-registered~\citep{nosek2018prereg}, following recommendations for rigorous evaluation in reinforcement and robot learning~\citep{henderson2018matters,agarwal2021precipice,kressgazit2024empirical}.

\section{A denoised counterfactual protocol}\label{sec:protocol}

\begin{figure*}[t]
\centering
\includegraphics[width=\linewidth]{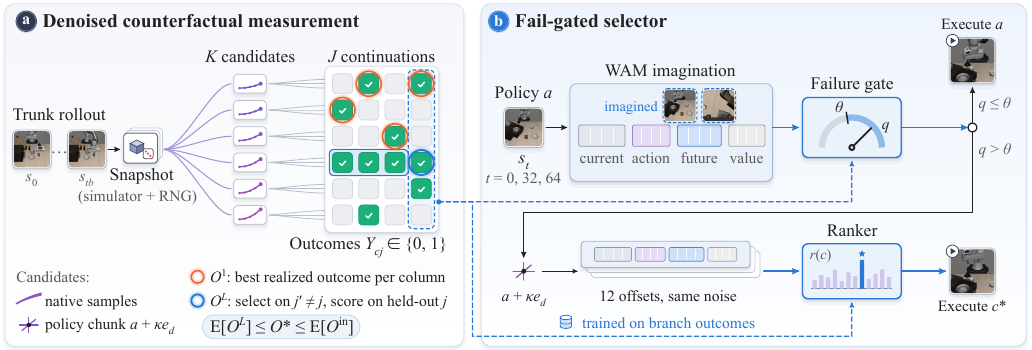}
\caption{\textbf{Overview.} (a)~From a bitwise-reproducible snapshot at $s_{t_b}$, each of $K$ candidate chunks is executed and continued $J$ times by the policy. $O^{1}$ (orange) takes the best realized outcome in each column; $O^{L}$ (blue) selects on the other columns and scores on the held-out one. Here every column contains a success, yet only one candidate succeeds in every continuation. (b)~At steps 0, 32, and 64, a failure gate reads the representation of the imagined policy chunk $a$; only when its score $q$ exceeds $\theta$ are twelve axis offsets of $a$ imagined with the same noise and ranked. Both readouts are trained on branch outcomes (dashed) rather than taken from the value head; the frames in (b) come from the episode of \cref{fig:closedloop}.}
\label{fig:overview}
\end{figure*}

\paragraph{Selectable headroom.} At a decision point $s_{t_b}$, the policy outputs an action chunk and executes its first part (16 steps for Cosmos Policy) before it is queried again. For a candidate $c$, let $p_c$ denote the probability of task success when the executed part of $c$ is applied from $s_{t_b}$ and the policy acts thereafter; the randomness is that of all later policy queries. The selectable headroom over a uniform choice among $K$ candidates is
\begin{equation}
O^\star=\max_c p_c-\frac{1}{K}\sum_c p_c .
\label{eq:ostar}
\end{equation}
When the pool contains the policy chunk $c=0$, we also report $\max_c p_c-p_0$, the gain over executing that chunk.

\paragraph{Bitwise-reproducible branching.} We snapshot and restore the full MuJoCo~\citep{todorov2012mujoco} state, the global random-number state, and the fixture states that RoboCasa keeps outside the joint positions (\eg, stove flames). Branching from a snapshot reproduces seeded replay from reset byte for byte (actions, proprioception, camera frames, and success flags), and every rollout passes online and offline rendering checks (\cref{app:protocol}).

\paragraph{Repeated continuations and held-out oracles.} Let $Y_{cj}\in\{0,1\}$ denote the outcome of candidate $c$ under continuation $j$, and let $\phat_c=\frac{1}{J}\sum_jY_{cj}$. We compare three estimators:
\begin{align}
O^{1}&=\tfrac{1}{J}\textstyle\sum_j\big[\max_cY_{cj}-\mathrm{mean}_cY_{cj}\big],\label{eq:o1}\\
O^{\mathrm{in}}&=\max_c\phat_c-\mathrm{mean}_c\phat_c,\label{eq:o4}\\
O^{L}&=\tfrac{1}{J}\textstyle\sum_j\big[\mathrm{mean}_{c\in B_j}Y_{cj}-\mathrm{mean}_cY_{cj}\big],\label{eq:ol}
\end{align}
where $B_j=\arg\max_c\mathrm{mean}_{j'\neq j}Y_{cj'}$, with ties averaged. $O^1$ is the standard single-continuation oracle. It depends only on whether each column contains a success, so it is invariant to shuffling candidate labels within a column and cannot distinguish candidate quality from continuation luck. $O^L$ selects with continuations that are independent of the continuation on which it is scored, so it measures what a selector with $J-1$ outcomes per candidate realizes, and $\mathbb E[O^L]\le O^\star\le\mathbb E[O^{\mathrm{in}}]$, where $O^{\mathrm{in}}$ is the in-sample oracle (\cref{app:estimators}). We report the weighted means $\Oone$, $\Oin$, and $\OL$ over decision points and the held-out share $R^L=\OL/\Oone$.

\paragraph{Variance decomposition.} Continuations are either independent across candidates or share one seed schedule, as in the protocol of \citet{yuan2026beyond}. With shared schedules, the two-way model
\begin{equation}
Y_{cj}=\mu+a_c+b_j+e_{cj}
\label{eq:anova}
\end{equation}
separates the candidate effect $\tau^2=\mathrm{Var}(a_c)$ from the schedule effect and from the candidate$\times$continuation interaction $e_{cj}$, which no selector can exploit. For structured proposals, we further split $\tauhat$ into a direction main effect shared by all states and a state-by-direction interaction.

\paragraph{Efficient sampling.} At most decision points, all candidates have identical outcomes. Each candidate therefore first receives one continuation; decision points with differing outcomes, together with a hash-selected quarter of the remaining points (weight 4), receive further continuations. Estimates use only the second-stage continuations, weighted by the inverse of the sampling probability, and confidence intervals (CIs) come from a bootstrap~\citep{efron1993bootstrap} clustered by state.

\paragraph{Proposals.} \emph{Native} candidates are independent samples from the policy. \emph{Structured} candidates displace the policy chunk $a$ along the six axes $e_d$ of the robot base,
\begin{equation}
a'_{k,0:3}=\mathrm{clip}\big(a_{k,0:3}+\kappa e_d,\,-1,\,1\big),
\label{eq:offset}
\end{equation}
with orientation and gripper unchanged; $\kappa=0.1$, $0.25$, and $0.5$ move the end effector by about 2, 5, and 10\,cm. Together with $a$, this yields seven candidates per scale.

\paragraph{Reading the imagination.} Value heads score either the real state after the action or the \emph{imagined} path used at deployment: the policy is queried at $s_{t_b}$, the candidate chunk is injected, the world model generates the future, and the value head scores it. To test whether the frozen representation encodes which candidate succeeds, we also fit linear probes~\citep{alain2016probes} that predict the within-decision outcome $\phat_c$ from the tokens of the final transformer block at the future-frame positions, centered within each decision point. The probes are weighted dual ridge regressions trained leave-one-state-out, and their predictions are residualized on the global direction estimated from the other states, so that a probe receives no credit for a direction that is good everywhere. The statistic is the within-decision Spearman correlation $\rho$, and its null distribution is obtained by permuting outcomes within decision points and refitting the whole pipeline; $p$-values count the observed statistic as one of the permutations~\citep{phipson2010permutation}. All evaluator seeds are independent of the candidate.

\section{Experimental setup}\label{sec:setup}

\paragraph{Models and benchmarks.} On RoboCasa, we use the public RoboCasa checkpoint of Cosmos Policy (2B parameters, 67.1\% average success), which is fine-tuned from a Cosmos video world model~\citep{nvidia2025cosmos}. Its public value head was trained only on demonstration replays; following the planning recipe of the original work, we fine-tune it on 1{,}200 policy rollouts and use the 15K-step model throughout. On LIBERO-Plus~\citep{fei2025liberoplus}, a perturbed version of LIBERO~\citep{liu2023libero} that leaves room for improvement, we use the released LIBERO checkpoints of Cosmos Policy, Flex-$\pi$~\citep{yan2026flexpi}, Fast-WAM~\citep{yuan2026fastwam}, and Streaming-WAM~\citep{huang2026streamingwam}, each reproduced on LIBERO-10 within the CI of its published success rate. Following a rule fixed before any branching data were collected, we evaluate on the two perturbation categories on which every model succeeds 30--90\% of the time, namely Objects Layout and Robot Initial States. All four WAMs enter the headroom measurements, the two whose released checkpoints imagine the consequences of given actions (Cosmos Policy and Streaming-WAM, the latter from the first 8 executed steps; \cref{app:protocol}) enter the imagination probes, and the selector is built on Cosmos Policy.

\paragraph{Data and pre-registration.} We collect twelve branching datasets, four on RoboCasa and eight on LIBERO-Plus (\cref{tab:data} in \cref{app:protocol}). Each dataset uses its own seed set, and initial-state hashes confirm that the datasets of each model do not overlap with one another or with the evaluation and training episodes. Every confirmatory test in the main text is pre-registered: its data, statistic, and criterion were registered before these data were collected (\cref{app:prereg} lists all 24 tests in the order of registration). Sections numbered with letters are in \suppref.

\section{Results}\label{sec:results}

\subsection{The single-continuation oracle is dominated by continuation randomness}\label{sec:noise}

\begin{figure}[t]
\centering
\includegraphics[width=\linewidth]{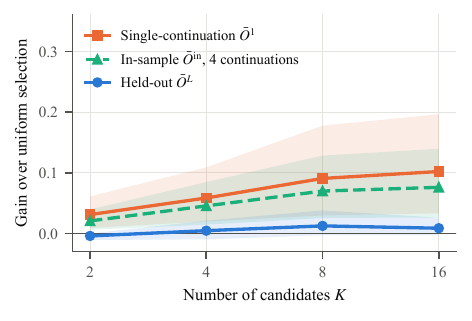}
\caption{\textbf{Headroom versus the number of candidates under the protocol of \citet{yuan2026beyond}} (three RoboCasa tasks, 16 native candidates, continuation seeds shared across candidates). The single-continuation oracle keeps rising with $K$, whereas the held-out oracle stays near zero. Shading: 95\% CIs.}
\label{fig:noise}
\end{figure}

We first follow the protocol of \citet{yuan2026beyond}: 16 native candidates per decision point, the first four re-planning points of each trajectory, and continuation seeds shared by all candidates, but with four additional seed schedules per decision point (three RoboCasa tasks; E3 in \cref{tab:data}). Using only the first schedule, we reproduce their observation: the best realized candidate raises success from 46.1\% to 58.3\% across 84 decision points (+12.3 points, 95\% CI $[3.9,22.5]$; they report +10.3 points).

Repeated continuations reveal where this gain comes from. At $K=16$, the single-continuation oracle is $\Oone=0.102\,[0.036,0.190]$, whereas the held-out oracle is $\OL=0.009\,[-0.001,0.023]$: a selector that observes three independent outcomes per candidate realizes $R^L=0.086\,[-0.009,0.270]$ of the measured gain on a fresh continuation, which meets the pre-registered criterion of an upper confidence bound below 0.5. As $K$ grows, $\Oone$ keeps rising while $\OL$ stays flat (\cref{fig:noise}), which is the signature of selection on noise. The variance decomposition leads to the same conclusion: the candidate effect is $\tauhat=0.0006$, whereas the candidate$\times$continuation interaction is 0.028, so 98\% of the outcome variation that distinguishes candidates is re-drawn whenever the continuation randomness changes. The deployable selector, the fine-tuned value head on the imagined future of each candidate, realizes $+0.006$, or 6\% of $\Oone$.

The same structure appears on LIBERO-Plus for all four WAMs (80 shared states; E5). Native candidates are again nearly identical (median end-effector distance 0.07--0.18\,cm), and the held-out oracle realizes 15\%, 18\%, and 9\% of the single-continuation gain for Cosmos Policy, Flex-$\pi$, and Fast-WAM, respectively, or $0.13\,[0.02,0.26]$ when the three models are pooled, which again meets the pre-registered criterion. For Streaming-WAM, the held-out oracle realizes none of its 3.1-point single-continuation gain ($R^L=-0.09\,[-0.23,0.05]$, pre-registered). For every model, the candidate effect is an order of magnitude smaller than the interaction (\cref{app:noise}).

\paragraph{Bracketing the selectable share.} Three binary outcomes per candidate form a noisy basis for selection, so $\OL$ is, in expectation, a lower bound on the selectable headroom. To bound it from both sides, we extended every second-stage decision point of Cosmos Policy, Flex-$\pi$, and Fast-WAM on LIBERO-Plus to 17 continuation schedules per candidate (\cref{fig:precision} in \suppref). For Cosmos Policy, the held-out share grows from 10\% with one selection schedule to 24\% $[9,40]$ with 16 and then flattens, whereas the in-sample oracle, whose expectation is an upper bound, reaches 52\% $[43,65]$; the two bounds are 33\% $[15,50]$ and 66\% $[55,79]$ for Flex-$\pi$ and 33\% $[15,51]$ and 63\% $[50,79]$ for Fast-WAM. At 95\% confidence, the single-continuation oracle thus overstates the selectable share of native headroom at least 1.5-fold for Cosmos Policy and 1.27-fold for the other two; in absolute terms, the upper bounds are 4.1, 2.8, and 4.0 points, against single-continuation estimates of 7.9, 4.3, and 6.4. Two alternative explanations do not hold. The estimator does not limit native candidates: on structured candidates, whose differences are real, three outcomes per candidate already recover 67--95\% (\cref{fig:teaser}b). Nor does a judge-free selector find native headroom: offline, KeyStone~\citep{dai2026keystone} gains $-0.004\,[-0.012,0.004]$ over the candidate mean on LIBERO-Plus and $+0.006\,[-0.002,0.015]$ on RoboCasa (\cref{app:noise}).

\textbf{Takeaway.} The single-continuation oracle measures how often the continuation of \emph{some} candidate succeeds, not how much better any candidate is; the two coincide only when continuations are deterministic. Held-out estimates, which require only one extra continuation per candidate, should accompany it whenever test-time selection is evaluated.

\subsection{Structured proposals create selectable headroom}\label{sec:proposals}

Native candidates are nearly identical: after 16 steps, their end-effector positions differ by a median of 0.15--0.35\,cm, and with independent continuations the held-out oracle over the policy chunk is $+0.003\,[-0.024,0.035]$ on TurnOnStove, even with increased sampling variance.

Displacing the policy chunk along the robot axes changes this picture (\cref{fig:teaser}b; \cref{app:proposals}). On TurnOnStove, the candidate effect grows from $\tauhat=0.027$ at $\kappa=0.1$ to 0.070 and 0.096 at $\kappa=0.25$ and 0.5, and the held-out oracle exceeds the policy chunk by $+0.158\,[0.081,0.247]$ at $\kappa=0.25$. At $\kappa=0.5$, the gains over the policy chunk are $+0.112$ on TurnOnStove, $+0.123\,[0.040,0.229]$ on TurnOnMicrowave, and $+0.131\,[0.049,0.238]$ on PnPCounterToSink. On LIBERO-Plus, with the perturbation scale matched to RoboCasa, the held-out oracle exceeds the policy chunk by $+0.112\,[0.046,0.196]$ for Cosmos Policy, $+0.088\,[0.034,0.157]$ for Fast-WAM, and $+0.045\,[0.006,0.096]$ for Flex-$\pi$ (all pre-registered). For Streaming-WAM, whose policy chunk succeeds most often (85\% at these decision points, against 66--75\%), the gain is $+0.045\,[-0.010,0.116]$ on the shared states and $+0.052\,[0.026,0.083]$ on 192 new states with three times as many decision points (E11; both pre-registered). This headroom is genuine: 67--95\% of the single-continuation gain survives held-out evaluation on both benchmarks and for all four WAMs, compared with at most 22\% for native candidates.

Three properties of this headroom determine what a selector must do. Perturbations spread outcomes in both directions: at $\kappa=0.5$, the average perturbed candidate is 7--14 points below the policy chunk, so the gain goes to a selector that finds the right candidate. The gain also lies where the policy chunk fails: in all four WAMs, the held-out oracle exceeds the chunk by 0.21--0.44 wherever the chunk fails in a selection continuation and by $-0.02$ to $+0.01$ elsewhere (exploratory; \cref{app:proposals}), so a stronger policy leaves a smaller average margin. The right direction also depends on the state: on TurnOnStove, the direction main effect accounts for only 2--6\% of $\tauhat$, and a fixed direction chosen on other states does not improve on the policy chunk. Structured proposals thus create headroom that a selector realizes through state-dependent choices at states where the policy is likely to fail, which is the design target of \cref{sec:selector}.

\subsection{The imagination carries the signal that the value head misses}\label{sec:imagination}

\begin{figure}[t]
\centering
\includegraphics[width=\linewidth]{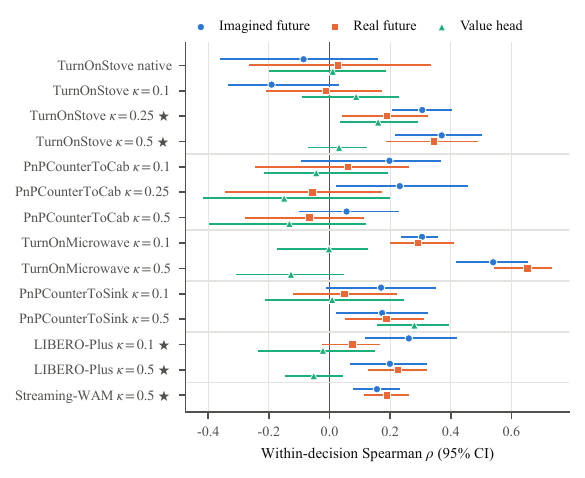}
\caption{\textbf{Candidate-outcome information in imagined and real futures.} Each row is a task and a proposal. Blue and orange: linear probes of the frozen representation of Cosmos Policy (last row: Streaming-WAM on 192 new states) for the imagined and the real future (RoboCasa: $s_{t_b+32}$; LIBERO-Plus: the state after the executed chunk), with the global direction removed. Green: the value head without additional training. $\star$: pre-registered test.}
\label{fig:imagination}
\end{figure}

\paragraph{Value heads value states.} Fine-tuning on policy rollouts makes the value head an accurate judge of states: the AUROC for final success rises from 0.53 to 0.87--0.90, and the Spearman correlation with Monte-Carlo returns rises from $-0.06$ to 0.68--0.80. Within a decision point, however, the ranking of candidates by the value head is essentially unrelated to their outcomes ($|\rho|\le0.05$ for native candidates), and common random numbers make this ranking stable across seeds but not informative; with independent seeds per candidate, as in deployed BoN, seed-to-seed noise dominates (\cref{app:imagination}).

\paragraph{The imagined future carries outcome information.} \Cref{fig:imagination} asks whether information that ranks candidates exists in the model at all. On TurnOnStove, a linear probe of the imagined future ranks perturbed candidates with $\rho=+0.31$ ($p=0.001$) at $\kappa=0.25$ and $+0.37$ ($p=0.0005$) at $\kappa=0.5$, which passes our pre-registered test; on TurnOnMicrowave, it reaches $+0.54$. On LIBERO-Plus, the probe for Cosmos Policy reaches $\rho=0.26$ ($p=0.020$) at $\kappa=0.1$ and $0.20$ ($p=0.004$) at $\kappa=0.5$, which again passes the pre-registered test. On Streaming-WAM, the probe reaches $\rho=0.16\,[0.08,0.23]$ ($p=0.0015$) on 192 new states, which passes the pre-registered test (0.11, $p=0.10$, on the 59 decision points of the shared states; \cref{app:imagination}). We detect no loss in the imagined future relative to the real one: across the ten RoboCasa perturbation conditions, the paired difference between the probes of imagined and real futures is $+0.02\pm0.05$ (exploratory), and on LIBERO-Plus the pre-registered equivalence test bounds it within $\pm0.15$ for Cosmos Policy ($+0.04\,[-0.06,0.13]$) and for Streaming-WAM on both state sets ($-0.005\,[-0.146,0.129]$ and $-0.033\,[-0.091,0.024]$). Averaged over the eleven RoboCasa conditions, the probe of the imagined future reaches $\rho=0.19$, the probe of the real future 0.15, and the value head 0.01. Qualitatively, the imagination reproduces the consequences that distinguish sibling candidates, including those of the candidates that fail (\cref{app:imagination}).

\paragraph{The predicted consequence, not only the action, carries the signal.} Because the candidate action is injected into the imagination pass, a probe of the imagined future could read the action rather than its consequence; the latent layout separates the two, because the action frame encodes the candidate action in context and the future frames hold the predicted consequence. Pooled over twelve perturbation conditions on both benchmarks (360 decision points), the future-frame probe reaches $\rho=0.23\,[0.18,0.28]$ and the action-frame probe $0.15\,[0.09,0.20]$. After removing what the action-frame probe explains, the future frames retain $0.11\,[0.05,0.17]$, whereas the action frame retains only $0.02\,[-0.04,0.07]$ beyond the future frames. The information even reaches the value frame, whose tokens yield $\rho=0.11\,[0.06,0.16]$, whereas the scalar output of the value head yields 0.01: the final projection of the evaluator discards what its own representation holds.

\textbf{Takeaway.} Within the pre-registered equivalence margin of $\pm0.15$, a linear readout of the final-block tokens of the imagined future matches the same readout of the real future for both WAMs over the predicted horizon, and for Cosmos Policy the information resides in the predicted consequence. The component that falls short is the value head: its state-value objective is dominated by differences between states rather than between sibling actions, and within the predicted horizon the information available for selection is lost in its final projection, not in the predicted frames (\cref{app:imagination}).

\subsection{A selector built on the findings improves on the policy}\label{sec:selector}

\begin{table}[t]
\centering\footnotesize
\setlength{\tabcolsep}{2.4pt}
\caption{\textbf{Fail-gated selector versus the policy} (Cosmos Policy on LIBERO-Plus; differences in points with 95\% CIs). Top: gain per decision point on 400 new states (the gate fires at 45\% of them). Bottom: closed-loop task success (three replicates per episode, common random numbers) on reserved episodes of the training tasks and of tasks absent from the training of the selector (the WAM checkpoint is the same). $^\dagger$Pre-registered test.}
\label{tab:selector}
\begin{tabular}{@{}lccc@{}}
\toprule
Single decision points (E9) & & & Gain \\
\midrule
Gate $+$ random offset & & & $-1.9\,[-3.5,-0.3]$ \\
Gate $+$ best fixed direction$^\dagger$ & & & $+0.1\,[-1.0,1.3]$ \\
Gate $+$ learned ranker$^\dagger$ & & & $+2.0\,[0.7,3.4]$ \\
\midrule
Closed loop (E8, E10) & Policy & $+$Selector & Difference \\
\midrule
Training tasks, all episodes$^\dagger$ & 76.9 & 81.6 & $+4.7\,[1.9,7.6]$ \\
Training tasks, robot initial states & 65.1 & 75.8 & $+10.7\,[5.4,16.3]$ \\
Unseen tasks$^\dagger$ & 62.6 & 69.8 & $+7.2\,[4.7,9.8]$ \\
Unseen tasks, trained on all tasks & 62.6 & 76.0 & $+13.4\,[10.7,16.2]$ \\
\bottomrule
\end{tabular}
\end{table}

\paragraph{Design.} The findings indicate where a selector can win. For Cosmos Policy on LIBERO-Plus, $+0.104$ of the $+0.112$ held-out oracle comes from decision points at which the policy chunk fails in all four continuations, where even a random choice among the seven candidates gains 16 points. We therefore split selection into two decisions (\cref{fig:overview}b). A \emph{failure gate}, a logistic readout of the representation that the WAM computes when imagining its own chunk, decides whether to deviate from the chunk; this is a state-level judgment of the kind that learned value heads make well, and the frames of the current scene alone match the full representation for this decision (\cref{app:selectors}). World-Coherent Decoding~\citep{zhang2026wcd} and adaptive execution~\citep{wang2026trust} act on the mismatch between imagined and observed frames; the gate is instead trained on the task outcomes of counterfactual branches to predict when the policy chunk will fail. A \emph{ranker}, the ridge readout of \cref{sec:imagination} applied to the imagination passes of twelve axis offsets ($\kappa\in\{0.1,0.5\}$), decides where to go and is consulted only when the gate fires.

\paragraph{Single-decision test.} On 400 new states (1{,}176 decision points; E9), the frozen selector improves on the policy chunk by $+2.0\,[0.7,3.4]$ points per decision point (\cref{tab:selector}). The gain comes from rescues: the selector deviates at 45\% of decision points, where the policy chunk succeeds only 52\% of the time, and where the policy chunk fails in all eight continuations, the chosen candidate succeeds 15\% of the time. Behind the same gate, a random offset loses 1.9 points and the offset that performs best on the training set gains nothing, so the ranker contributes state-dependent choices rather than a fixed correction (both comparisons pre-registered). Within the training tasks, a ranker that reads only the current state and the commanded direction comes close to the full model in nested cross-validation ($+1.9$ against $+2.5$ points; \cref{app:selectors}), so ranking there does not require the imagination. All features, regularization strengths, and thresholds were chosen on a training set (320 LIBERO-Plus states; E7) from a pre-declared grid by cross-validation over states (nested cross-validation estimates the gain of this procedure at $+2.5\,[0.9,4.1]$ points), and the model was frozen before any test data existed.

\paragraph{Closed loop.} In closed loop on reserved episodes of the training tasks, the frozen selector raises task success from 76.9\% to 81.6\% ($+4.7\,[1.9,7.6]$ points, pre-registered). The gain is largest where the policy fails most often, which is also where the structured proposals place the headroom: under robot initial-state perturbations, success rises from 65.1\% to 75.8\% ($+10.7\,[5.4,16.3]$ points), and the held-out oracle there exceeds the policy chunk by $+0.222$, against $+0.031$ under object-layout perturbations, where the policy already succeeds 88.6\% of the time (\cref{app:selectors}). The 480 episodes (E8) were fixed before any selector was trained and unread until this test; each runs three times per arm with reset and query seeds shared by both arms, so the 576 episodes without deviation are identical to their policy counterparts, as verified. Because offsets are imagined only when the gate fires, the selector adds 17.5 imagination passes per episode to 12 policy queries; offline, this gate attains nearly the largest gain while imagining offsets at a third of the decision points, whereas imagining them at every decision point gains a third as much (\cref{fig:cost}).

\begin{figure}[t]
\centering
\includegraphics[width=\linewidth]{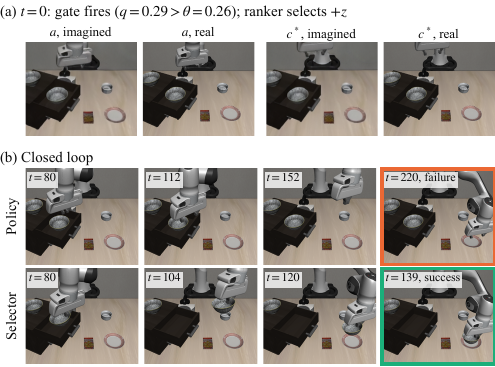}
\caption{\textbf{Closed-loop example on an unseen task} (both runs share all seeds). (a)~At $t=0$, the gate fires on the policy chunk $a$, and the ranker selects $c^{*}$, the upward offset with $\kappa=0.5$; imagined frames next to the real frames 16 steps later. (b)~The policy does not lift the bowl from the drawer and times out; the selector succeeds without deviating again. The episode is the first, in hash order, of the 62 in which the policy fails and the selector succeeds in all three replicates; the first of the 26 reverse cases is in \cref{fig:closedloop_harm} (\suppref).}
\label{fig:closedloop}
\end{figure}

\paragraph{Unseen tasks.} The selector also transfers to tasks absent from its training. Success rises from 62.6\% to 69.8\% with a selector that never saw the task ($+7.2\,[4.7,9.8]$ points, pre-registered; $[3.2,11.1]$ when tasks are resampled) and to 76.0\% with the selector trained on all tasks. The selector is re-trained without the tasks of each of four suite-stratified folds of the 40 base tasks, with cross-validation grouped by task, and frozen before the test, which runs all 858 unused robot initial-state episodes (E10) three times with the policy and with both selectors. The gain is positive in every fold and in 30 of the 40 tasks (\cref{fig:e10tasks}; example in \cref{fig:closedloop}), so about half of the closed-loop gain carries over to unseen tasks.

\textbf{Takeaway.} Splitting the decision into whether to deviate and where to go turns measured headroom into pre-registered improvements over the policy. The gate spends the additional imagination where the policy is likely to fail, so the gain is largest where the policy is weakest.

\section{HeadroomBench}\label{sec:bench}

Our measurements are inexpensive to repeat, and their branching data are exactly what a selector needs for training, so we package both as HeadroomBench for public release. For any policy with seeded queries and any simulator with state snapshots, the headroom track measures the single-continuation oracle, the held-out share, the headroom over the policy chunk, and the variance decomposition, with state-clustered CIs, at a cost close to that of the standard protocol. The selector track provides a training split, a test split, and closed-loop evaluation lists, with reference results for value heads, random offsets, and both of our selectors (\cref{app:bench}). On its test split, the held-out oracle exceeds the policy chunk by 11 points, whereas selection by the value head and random selection both lose 8--10 points, which leaves most of the headroom open.

\section{Discussion and conclusion}\label{sec:conclusion}

Test-time selection can improve a world action model only as much as its selectable headroom allows. The single-continuation oracle measures how often the continuation of some candidate succeeds, whereas selection depends on how much better the best candidate is, which the held-out oracle measures at the cost of one extra continuation per candidate. Across four WAMs and two benchmarks, we find that the standard oracle largely reflects continuation randomness, that real headroom appears when proposals make consequential changes at states where the policy fails, and that information that helps find this headroom is already present in the imagined futures of the model, with no detectable loss relative to the real future. A selector that deviates only when a learned readout predicts failure improves on the policy at fresh decision points and in closed loop, gains most where the policy fails most often, and carries about half of its closed-loop gain over to tasks absent from its training.

\paragraph{Scope.} Our measurements cover four WAMs on two MuJoCo-based benchmarks, where bitwise-reproducible branching is available; the protocol requires only state snapshots and seeded policy queries and extends to any WAM and simulator that provide them. On physical robots, where a state cannot be restored exactly, repeated resets approximate the held-out estimate up to reset noise, whereas the selector itself needs no state restoration at deployment, because it scores offsets with the imagination of the WAM. Advantage heads trained on sibling branches, the data that our protocol produces, are a natural next step toward the held-out oracle.

%% file: appendix.tex
\appendix
\renewcommand{\dblfloatpagefraction}{0.4}  %
\makeatletter  %
\setlength{\@fptop}{0pt}\setlength{\@fpsep}{\floatsep}\setlength{\@fpbot}{0pt plus 1fil}
\setlength{\@dblfptop}{0pt}\setlength{\@dblfpsep}{\dblfloatsep}\setlength{\@dblfpbot}{0pt plus 1fil}
\makeatother

\section{Pre-registration}\label{app:prereg}

\Cref{tab:prereg,tab:prereg2} list all confirmatory tests of the project with their outcomes, in the order of registration. A test is pre-registered when its data, statistic, and criterion were registered before its data were collected: every test was registered before the first rollout of the dataset that it introduced. The imagined-future probe on TurnOnStove was registered in the experiment plan before the perturbation-scale rollouts of E2. Ranking with abstention was registered before the rollouts of its training set E7, and thus before any selector was trained; its test split E6 had been collected for the preceding tests. The structured-headroom test for Fast-WAM was registered after the first ten tests, and the comparison with a fixed direction was registered after the outcome of the single-decision test of the fail-gated selector was known and before any rollout of the fixed-direction arm, with the fixed direction determined by E7 alone. The tests on Streaming-WAM and on 17 continuations for Flex-$\pi$ and Fast-WAM were registered together, after the first fifteen tests and before any of their data. The three tests on E11 were registered after the outcomes of the Streaming-WAM tests on E6 were known, to repeat them on new states with a larger sample, and before any E11 data; the wrist-camera feature of the E11 probe comes from an exploratory result on E6. The pre-registration texts and their repository history will be released with HeadroomBench.

\begin{table*}[p]
\centering\footnotesize
\caption{All pre-registered tests and their outcomes, in the order of registration.}
\label{tab:prereg}
\begin{tabular}{@{}p{2.5cm}p{6.5cm}p{7.4cm}@{}}
\toprule
Test & Hypothesis and criterion & Outcome \\
\midrule
Imagined-future probe, TurnOnStove (E2) & The imagined-future probe ranks perturbed candidates on TurnOnStove at $\kappa=0.25$ and $0.5$; criterion: permutation $p<0.025$ and CI lower bound $>0$ & Supported: $\rho=+0.31$ ($p=0.001$) and $+0.37$ ($p=0.0005$) \\
Held-out share, replication (E3) & Ratio $R^L$ of the held-out to the single-continuation oracle; premise: at least 15 weighted decision points and a CI lower bound of $\Oone$ above 0; criterion: upper CI bound $<0.5$ & Supported: $R^L=0.086\,[-0.009,0.270]$; secondary test $\OL>0$: $p=0.12$ \\
Onset of readability (E4) & The perturbation scale at which outcomes become readable on TurnOnStove (readable at $\kappa=0.5$ with $p<0.0125$, not at $\kappa=0.1$) transfers to TurnOnMicrowave and PnPCounterToSink & Not supported, because readability can appear earlier: TurnOnMicrowave is readable already at $\kappa=0.1$ ($\rho=+0.31$, $p=0.028$) and at $\kappa=0.5$ ($+0.54$, $p=0.0005$); PnPCounterToSink at $\kappa=0.5$: $\rho=+0.17$, $p=0.032$. The onset of readability depends on the task (\cref{app:imagination}) \\
Value head on real states (E4) & A value head that scores real post-action states selects better candidates than one that scores imagined futures & Not supported, consistent with \cref{sec:imagination}: value heads rank candidates poorly whatever their input (gain $+0.014$, $p=0.20$; paired difference $+0.001\,[-0.044,0.049]$) \\
Held-out share, three WAMs (E5) & $R^L$ pooled over the WAMs that meet the premise of the replication test; criterion: upper CI bound $<0.5$ & Supported: all three models meet the premise; pooled $R^L=0.134\,[0.024,0.260]$; per model 0.145, 0.177, and 0.092, each with an upper bound $<0.5$ \\
Structured headroom (E6) & Held-out oracle among seven candidates minus the policy chunk at $\kappa=0.5$, per model; criterion: CI lower bound $>0$ & Supported for both models: Cosmos Policy $+0.112\,[0.046,0.196]$; Flex-$\pi$ $+0.045\,[0.006,0.096]$ \\
Imagined-future probe, LIBERO-Plus (E6) & Imagined-future probe (specification of the TurnOnStove probe), per model; criterion: on some arm, permutation $p<0.025$ and CI lower bound $>0$ & Supported for Cosmos Policy: $\rho=0.262$ ($p=0.020$) at $\kappa=0.1$ and $0.199$ ($p=0.004$) at $\kappa=0.5$; not testable for Flex-$\pi$, whose released checkpoint has no action-conditioned generation \\
Imagined versus real probe (E6) & Imagined minus real probe, pooled over the arms that meet the premise of the LIBERO-Plus probe; criterion: CI within $[-0.15,0.15]$ & Supported: $+0.038\,[-0.057,0.132]$ \\
In-sample share, 17 continuations (E5$^{+}$) & In-sample oracle over 17 continuation schedules as a share of $\Oone$ (Cosmos Policy, 16 native candidates); criterion: upper CI bound $<0.5$ & Not supported: the in-sample share is $0.519\,[0.433,0.648]$, so the overstatement is at least 1.5-fold at 95\% confidence; the held-out share with 8--9 selection schedules is $0.224\,[0.089,0.367]$ (\cref{fig:precision}) \\
Ranking with abstention (E7$\to$E6) & Ridge selector on imagined-future tokens with an abstention threshold, hyperparameters chosen by cross-validation on E7; gain over the policy chunk on E6 at $\kappa=0.5$; criterion: CI lower bound $>0$ & Not supported: cross-validation chose never to deviate at every training size, so the selector matches the policy (gain $0\,[0,0]$) instead of losing like the value head ($-0.096$; held-out oracle $+0.112$); this outcome motivated the fail-gated selector (\cref{app:selectors}) \\
Structured headroom, Fast-WAM (E6) & As the structured-headroom test, for Fast-WAM (registered after the first ten tests); criterion: CI lower bound $>0$ & Supported: $+0.088\,[0.034,0.157]$; at $\kappa=0.1$: $+0.053\,[0.021,0.100]$ \\
Fail-gated selector, single decisions (E9) & Fail-gated selector, frozen after development on E7 and before this test, versus the policy chunk on 400 new states, with eight continuations per arm where it deviates; criterion: CI lower bound $>0$ & Supported: $+0.020\,[0.007,0.034]$; deviation at 45\% of decision points; the same gate with a random offset gives $-0.019\,[-0.035,-0.003]$ \\
Selector versus fixed direction (E9) & Selector versus the same gate with a fixed direction, namely the offset with the highest average outcome on E7 ($\kappa=0.1$, $-y$), with eight continuations per arm where the gate fires; criterion: CI lower bound $>0$ & Supported: $+0.019\,[0.005,0.033]$; the gate with the fixed direction versus the policy chunk gives $+0.001\,[-0.010,0.013]$ \\
Closed loop, training tasks (E8) & Closed-loop task success with the frozen selector (called at steps 0, 32, and 64) minus that of the policy alone, on 480 reserved episodes $\times$ 3 replicates with common random numbers; episode-clustered bootstrap; criterion: CI lower bound $>0$ & Supported: $+0.047\,[0.019,0.076]$ (76.9\% $\to$ 81.6\%); robot initial states $+0.107$, object layouts $-0.013$ \\
Closed loop, unseen tasks (E10) & Closed-loop task success with a selector trained without the test task minus that of the policy alone; 40 base tasks in four folds, with one selector per fold trained on E7 without the tasks of that fold; all 858 unused robot initial-state episodes $\times$ 3 replicates; episode-clustered bootstrap; criterion: CI lower bound $>0$ & Supported: $+0.072\,[0.047,0.097]$ (62.6\% $\to$ 69.8\%); task-clustered CI $[0.032,0.111]$; the selector trained on all tasks gives $+0.134$ \\
\bottomrule
\end{tabular}
\end{table*}

\begin{table*}[t]
\centering\footnotesize
\caption{Pre-registered tests that extend the measurements to Streaming-WAM and to 17 continuations for Flex-$\pi$ and Fast-WAM (continued from \cref{tab:prereg}).}
\label{tab:prereg2}
\begin{tabular}{@{}p{2.5cm}p{6.5cm}p{7.4cm}@{}}
\toprule
Test & Hypothesis and criterion & Outcome \\
\midrule
Structured headroom, Streaming-WAM (E6) & As the structured-headroom test, for Streaming-WAM; criterion: CI lower bound $>0$ & Not supported: $+0.045\,[-0.010,0.116]$; its policy chunk succeeds at 85\% of these decision points (66--75\% for the other models), and 78\% of the single-continuation gain survives held-out evaluation (\cref{app:proposals}) \\
Imagined-future probe, Streaming-WAM (E6) & Imagined-future probe (specification of the LIBERO-Plus probe; imagination conditioned on the first 8 executed steps); criterion: on some arm, permutation $p<0.025$ and CI lower bound $>0$ & Not supported: both arms meet the premise; $\rho=0.079$ ($p=0.32$) at $\kappa=0.1$ and $0.110\,[-0.014,0.243]$ ($p=0.10$) at $\kappa=0.5$; the probe of the real future reaches $0.162$ ($p=0.037$) at $\kappa=0.5$ (\cref{app:imagination}) \\
Imagined versus real probe, Streaming-WAM (E6) & As the imagined-versus-real test, for Streaming-WAM; criterion: CI within $[-0.15,0.15]$ & Supported: $-0.005\,[-0.146,0.129]$ \\
Held-out share, Streaming-WAM (E5) & $R^L$ for Streaming-WAM, with the premise of the replication test; criterion: upper CI bound $<0.5$ & Supported: $R^L=-0.087\,[-0.227,0.052]$ (53 decision points); pooled with the other three models, $0.095\,[-0.008,0.213]$ (descriptive) \\
In-sample share, Flex-$\pi$ (E5$^{+}$) & In-sample oracle over 17 continuation schedules as a share of $\Oone$ (Flex-$\pi$, 16 native candidates); criterion: upper CI bound $<2/3$, that is, an overstatement of at least 1.5-fold & Not supported: $0.658\,[0.546,0.787]$, so the overstatement is at least 1.27-fold at 95\% confidence; the held-out share with 16 selection schedules is $0.334\,[0.146,0.504]$ (\cref{fig:precision}) \\
In-sample share, Fast-WAM (E5$^{+}$) & As the previous test, for Fast-WAM & Not supported: $0.626\,[0.499,0.786]$, so the overstatement is at least 1.27-fold at 95\% confidence; the held-out share with 16 selection schedules is $0.331\,[0.148,0.506]$ \\
Structured headroom, Streaming-WAM, new states (E11) & As the structured-headroom test, on 192 new states at $\kappa=0.5$; criterion: CI lower bound $>0$ & Supported: $+0.052\,[0.026,0.083]$ (195 decision points, 146 states) \\
Imagined-future probe, Streaming-WAM, new states (E11) & Specification of the LIBERO-Plus probe at $\kappa=0.5$, with two features tested once each: the final-block tokens and the wrist-camera spatial mean of the future frame; criterion: for some feature, permutation $p<0.025$ and CI lower bound $>0$ & Supported for both features: $\rho=0.157\,[0.077,0.232]$ ($p=0.0015$) and $0.102\,[0.029,0.173]$ ($p=0.017$); the probes of the real future reach $0.190$ and $0.195$ \\
Imagined versus real probe, Streaming-WAM, new states (E11) & Imagined minus real probe for the final-block tokens; criterion: CI within $[-0.15,0.15]$ & Supported: $-0.033\,[-0.091,0.024]$; for the wrist-camera feature (descriptive), $-0.094\,[-0.178,-0.009]$ \\
\bottomrule
\end{tabular}
\end{table*}

\section{Datasets and protocol details}\label{app:protocol}

\Cref{fig:overview}a summarizes the protocol. After a candidate chunk has been executed (16 steps on RoboCasa), the resulting state is snapshotted again, so that all continuations of the candidate start from the same state.

\begin{table*}[t]
\centering\footnotesize
\caption{Branching datasets. ``States'' counts trunks; each trunk contributes decision points at the listed $t_b$.}
\label{tab:data}
\begin{tabular}{@{}p{0.6cm}p{4.6cm}p{1.8cm}p{4.0cm}p{2.9cm}p{1.0cm}@{}}
\toprule
Set & Tasks (states) & $t_b$ & Candidates & Continuations & Rollouts \\
\midrule
E1 & TurnOnStove, PnPCounterToCab (17) & 0/48/96/144 & 6 native; random offsets & independent, $J{=}4$ & 2.4k \\
E2 & TurnOnStove (40), PnPCounterToCab (20) & 0/96 & chunk $+$ 6 axis offsets, $\kappa\in\{0.1,0.25,0.5\}$; native & independent, $J{=}4$ & 9.3k \\
E3 & TurnOffStove, PnPCounterToSink, TurnOnStove (7 each) & 0/16/32/48 & 16 native & shared schedule, $J{=}1{+}4$ & 3.9k \\
E4 & TurnOnMicrowave, PnPCounterToSink (25 each) & 0/96 & chunk $+$ 6 axis offsets, $\kappa\in\{0.1,0.5\}$ & independent, $J{=}4$ & 5.6k \\
\midrule
E5 & LIBERO-Plus, 4 WAMs (80 shared) & first query, step $\ge$80 & 16 native & shared schedule, $J{=}1{+}4$ & 21.5k \\
E5$^{+}$ & LIBERO-Plus, Cosmos Policy, Flex-$\pi$, and Fast-WAM, all second-stage points of E5 (37, 40, 35) & as E5 & 16 native & shared schedule, $J{=}17$ & 26.1k \\
E6 & LIBERO-Plus, 4 WAMs (56 shared) & first query, step $\ge$80 & chunk $+$ 6 axis offsets, $\kappa\in\{0.1,0.5\}$ & independent, $J{=}4$ & 17.5k \\
E7 & LIBERO-Plus, Cosmos Policy (320; selector training) & 0/32/64 & chunk $+$ 12 axis offsets, $\kappa\in\{0.1,0.5\}$ & independent, $J{=}1{+}2$ & 30.4k \\
E8 & LIBERO-Plus, Cosmos Policy (480; closed loop) & 0/32/64 & policy versus selector, 3 replicates & closed-loop episodes & 2.9k \\
E9 & LIBERO-Plus, Cosmos Policy (400; selector test) & 0/32/64 & chunk, selected and random offset & independent, $J{=}8$ where deviating & 12.3k \\
E10 & LIBERO-Plus, Cosmos Policy (858; unseen tasks, 4 folds) & 0/32/64 & policy versus two selectors, 3 replicates & closed-loop episodes & 7.7k \\
E11 & LIBERO-Plus, Streaming-WAM (192 new) & as E5 & chunk $+$ 6 axis offsets, $\kappa=0.5$ & independent, $J{=}4$ & 7.8k \\
\bottomrule
\end{tabular}
\end{table*}

\paragraph{Determinism and rendering integrity.} Snapshot branching is validated against seeded replay at every decision point (the restored observations and proprioception equal those of the trunk), on spot-check episodes (full episodes are byte-identical, including JPEG frames), and by re-running episodes in fresh processes. Every rollout also passes online and offline rendering checks. These checks guard against a failure mode of robosuite~\citep{zhu2020robosuite} that we identified while reproducing the baseline: the previous rendering context can be released by the garbage collector in the middle of an episode, which deletes same-named frame buffers of the active context and corrupts all subsequent frames (20 of 3{,}600 baseline evaluation episodes were affected). We force garbage collection at fixed points and verify frames online by read-back and offline by a frame-integrity scan. The LIBERO stack releases rendering contexts synchronously at reset, and the same checks are applied there. Flagged rollouts were adjudicated with four criteria (the online check passed, the other cameras are continuous, visual inspection, and a byte-identical re-run in a fresh process). All of them were adjudicated as false positives caused by fast wrist-camera motion over high-contrast textures, except two rollouts in which the wrist camera turns dark while the gripper is inside a dark container (one in E7 and one in E9); both meet all four criteria and were retained as genuine occlusions, and removing them changes no reported result by more than 0.003.

\paragraph{LIBERO-Plus models.} The released LIBERO checkpoints of Cosmos Policy, Flex-$\pi$ (\texttt{flexpi-libero-fulljoint-star}), Fast-WAM (\texttt{libero\_uncond\_2cam224}), and Streaming-WAM (\texttt{ac-stream}) were reproduced on LIBERO-10 with their official evaluation scripts (50 episodes each: 96\%, 98\%, 94\%, and 100\%, all within the CI of the published values). Each model received a per-query seed interface that changes only the random seed; full episodes are byte-identical across fresh processes and concurrent workers, and snapshot branching is byte-identical to seeded replay for every model. Streaming-WAM is designed for asynchronous execution, in which each query is conditioned on the 8 actions already committed from the previous chunk. Our protocol queries a policy once and executes its chunk, so we issue every 16 steps the call that its official controller uses at the start of an episode, which takes no committed prefix; with this synchronous schedule, it succeeds in 50 of 50 LIBERO-10 episodes. To imagine a candidate, we use the call with a committed prefix, which accepts exactly 8 actions: the first 8 of the 16 executed steps of the candidate are supplied, and the model predicts the rest. Because axis offsets are constant along the chunk, this prefix fixes the direction and size of the offset. The probe reads the 98 final-block tokens of the video expert at the latent frame that covers steps 4--16, and real futures are encoded into the same positions with the VAE of the model. Imagined and real latents agree closely for the policy chunk (cosine 0.92) and less for offsets (0.75--0.80), consistent with the unseen second half of the chunk. Native sampling of Streaming-WAM is stochastic (ratio of the standard deviation of chunks across seeds 0.040, against 0.015--0.12 for the other models), and it succeeds 71\% and 88\% of the time on the two selected categories. RynnVLA-002~\citep{cen2025rynnvla002} decodes greedily in its official configuration, so its native candidates are identical; the study therefore uses the four WAMs above.

\paragraph{LIBERO-Plus environment.} Three properties of LIBERO matter for reproducible branching, and we handle them uniformly. The global random state is seeded before every reset; a fresh environment is created for every episode and branch, because repeated resets of one environment object append placement samplers that move articulated fixtures by up to 1.3\,cm; and snapshots store the trunk observation, because LIBERO-Plus adds sensor noise inside the \texttt{step} function of its wrapper. The task names of LIBERO-Plus carry perturbation suffixes; except in the language category, we use the original LIBERO instruction, and all success rates refer to these instructions.

\paragraph{Categories, states, and proposals.} A list of 168 episodes (4 suites $\times$ 7 categories $\times$ 6) was run with every model before any branching data were collected. The rule announced beforehand selects the categories on which all models succeed 30--90\% of the time, namely Objects Layout and Robot Initial States. E5 draws 80 states (10 per suite and category) shared by all models, and E6 draws 56 different states. E5$^{+}$ uses 17 continuation schedules at every second-stage decision point of E5, namely the first-stage schedule $j=0$, the four second-stage schedules $j=1..4$, and the new schedules $j=5..16$, all shared across candidates; its estimates average over these 17 schedules, whereas those of E5 use $j=1..4$, so the two estimates of $\Oone$ differ (for example, 7.9 against 7.3 points for Cosmos Policy). E11 draws 192 further states (24 per suite and category) for Streaming-WAM alone; they exclude F1, E5, E6, E8, and E10, and because the other robot initial-state episodes are all in E10, they may coincide with states of E7 and E9, which contain only Cosmos Policy (21 and 25 identical initial frames). Decision points are the first policy query and the first query at or after environment step 80. Native candidates execute the chunk length of each model (16 steps for Cosmos Policy and Streaming-WAM, 10 for Flex-$\pi$ and Fast-WAM). Structured offsets are added in the environment action space and scaled by $16/n_{\mathrm{exec}}$, so that the cumulative commanded displacement matches RoboCasa ($\kappa=0.1$ and $0.5$ move the end effector by about 2 and 6--10\,cm).

\section{Estimators}\label{app:estimators}

\paragraph{Permutation invariance of $O^1$.} For each continuation $j$, $\max_cY_{cj}$ and $\mathrm{mean}_cY_{cj}$ are symmetric functions of the column $\{Y_{cj}\}_c$. Permuting candidate labels independently within columns therefore leaves $O^1$ unchanged, so $O^1$ carries no information about which candidate is better.

\paragraph{Bounds.} By Jensen's inequality, $\mathbb E[\max_c\phat_c]\ge\max_c\mathbb E[\phat_c]=\max_cp_c$, so $\mathbb E[O^{\mathrm{in}}]\ge O^\star$. For $O^L$, the selected set $B_j$ is independent of column $j$, so $\mathbb E[\mathrm{mean}_{c\in B_j}Y_{cj}]=\mathbb E[\mathrm{mean}_{c\in B_j}p_c]\le\max_cp_c$, which gives $\mathbb E[O^L]\le O^\star$. With $K=16$, $J=4$, and no candidate effect at all, simulation gives $\mathbb E[O^{\mathrm{in}}]/\mathbb E[O^1]$ between 0.43 and 0.82 for base rates between 0.1 and 0.5, which is why $O^{\mathrm{in}}$ with four continuations is not used for testing.

\paragraph{Two-stage weights.} Decision points with differing first-stage outcomes are always continued (inclusion probability $\pi_i=1$, weight $w_i=1$); among the remaining points, a hash-selected quarter is continued ($\pi_i=1/4$, $w_i=4$). Every reported mean of a per-decision statistic $Z_i$ is the self-normalized weighted mean $\sum_iw_iZ_i/\sum_iw_i$ over the second-stage decision points. Because every decision point enters with an expected weight of 1, the numerator and the denominator are unbiased for the population total and the population size, so the mean is a consistent ratio estimator of the population mean; ratios such as $R^L$ divide two such means. The state-cluster bootstrap resamples states and recomputes numerators and denominators together.

\paragraph{Variance decomposition.} Under \cref{eq:anova}, let $\mathrm{MS}_{\mathrm{res}}$ denote the residual mean square of the two-way table. Then $\tauhat=S^2_c(\phat_c)-\mathrm{MS}_{\mathrm{res}}/J$, the schedule effect is $S^2_j(\bar Y_{\cdot j})-\mathrm{MS}_{\mathrm{res}}/K$, and $\mathrm{MS}_{\mathrm{res}}$ estimates the interaction, including binomial noise.

\paragraph{Model-based estimate of the selectable share.} Let candidate means differ by $a_c$ with variance $\tau^2$, and let the candidate$\times$continuation interaction, including binomial noise, have variance $\sigma^2$. Selecting on $m$ shared schedules ranks candidates by $a_c$ plus noise of variance $\sigma^2/m$, and under a Gaussian approximation the held-out gain is
\begin{equation}
G(m)=\frac{O^\star}{\sqrt{1+\lambda/m}},\qquad \lambda=\sigma^2/\tau^2 .
\label{eq:gaussian}
\end{equation}
On E5$^{+}$, the variance decomposition gives $\lambda=\mathrm{MS}_{\mathrm{res}}/\tauhat=10.5$, so $O^\star=G(16)\sqrt{1+\lambda/16}$, which is 31\% $[14,49]$ of $\Oone$ (bootstrap over states). Fitting $(O^\star,\lambda)$ directly to the curve $G(1),\dots,G(16)$ gives a close value, 27\% with $\lambda=5.0$. For Flex-$\pi$ and Fast-WAM, the moment version gives 46\% $[26,71]$ with $\lambda=13.6$ and 45\% $[23,78]$ with $\lambda=14.1$. Because the approximation ignores the heterogeneity of $\tau^2$ across decision points, we report it as exploratory.

\begin{figure*}[t]
\centering
\includegraphics[width=\linewidth]{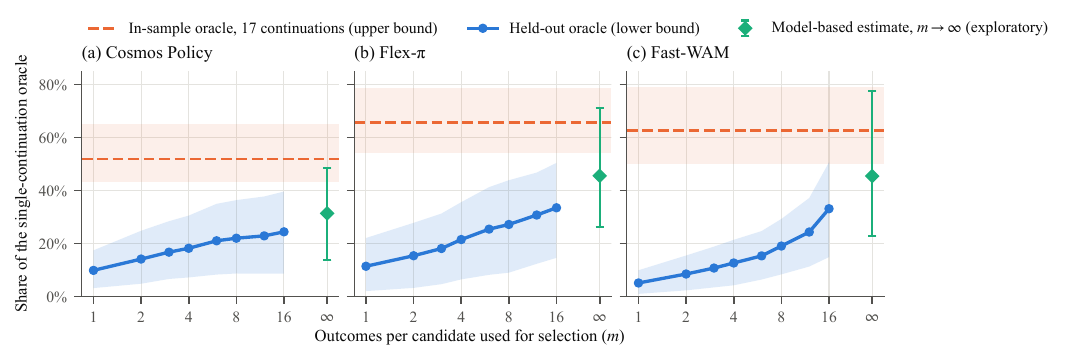}
\caption{\textbf{Bracketing the selectable share} (E5$^{+}$: Cosmos Policy (a), Flex-$\pi$ (b), and Fast-WAM (c) on LIBERO-Plus, 16 native candidates, 17 shared continuation schedules). Blue: held-out oracle that selects on $m$ schedules and is scored on another, as a share of the single-continuation oracle; its expectation is a lower bound. Orange: in-sample oracle over all 17 schedules; its expectation is an upper bound. Green: model-based extrapolation to $m\to\infty$ (exploratory). Bands and bars: 95\% CIs.}
\label{fig:precision}
\end{figure*}

\paragraph{Effective sample size and stability.} Two-stage weighting concentrates each ratio on the decision points that enter the second stage, and the single-continuation oracle concentrates on the subset of those points with mixed outcomes. \Cref{tab:stability} reports these counts and the range of each ratio when any one state is left out. Both ends of the bracket in \cref{fig:precision} are stable: leaving out any single state keeps them within $[21,30]\%$ and $[50,55]\%$ for Cosmos Policy within $[28,37]\%$ and $[64,71]\%$ for Flex-$\pi$, and within $[29,37]\%$ and $[59,67]\%$ for Fast-WAM.

\begin{table}[t]
\centering\footnotesize
\setlength{\tabcolsep}{1.6pt}
\caption{Effective sample size and leave-one-state-out (LOSO) stability of the headline ratios. Top-5: share of the weighted $\Oone$ carried by the five largest decision points. For the pooled row, States counts model--state pairs, and LOSO leaves out one such pair; the CIs of pooled estimates in the main text instead resample shared states.}
\label{tab:stability}
\resizebox{\linewidth}{!}{\begin{tabular}{@{}lcccccc@{}}
\toprule
Ratio & DPs & States & $O^1>0$ & Top-5 & Est. & LOSO range \\
\midrule
E3 $R^L$ & 36 & 18 & 19 & 0.43 & 0.086 & $[0.030,0.130]$ \\
E5 $R^L$, Cosmos & 46 & 37 & 21 & 0.49 & 0.145 & $[0.112,0.194]$ \\
E5 $R^L$, Flex-$\pi$ & 46 & 40 & 18 & 0.61 & 0.177 & $[0.062,0.246]$ \\
E5 $R^L$, Fast-WAM & 44 & 35 & 20 & 0.59 & 0.092 & $[0.033,0.173]$ \\
E5 $R^L$, Streaming-WAM & 53 & 41 & 16 & 0.66 & $-0.087$ & $[-0.119,-0.036]$ \\
E5 $R^L$, pooled (first three) & 136 & 112 & 59 & 0.27 & 0.134 & $[0.107,0.162]$ \\
E5$^{+}$ held-out, Cosmos & 46 & 37 & 22 & 0.51 & 0.244 & $[0.207,0.296]$ \\
E5$^{+}$ in-sample, Cosmos & 46 & 37 & 22 & 0.51 & 0.519 & $[0.499,0.547]$ \\
E5$^{+}$ held-out, Flex-$\pi$ & 46 & 40 & 21 & 0.57 & 0.334 & $[0.276,0.371]$ \\
E5$^{+}$ in-sample, Flex-$\pi$ & 46 & 40 & 21 & 0.57 & 0.658 & $[0.638,0.705]$ \\
E5$^{+}$ held-out, Fast-WAM & 44 & 35 & 24 & 0.46 & 0.331 & $[0.285,0.370]$ \\
E5$^{+}$ in-sample, Fast-WAM & 44 & 35 & 24 & 0.46 & 0.626 & $[0.594,0.666]$ \\
\bottomrule
\end{tabular}}
\end{table}

\section{Additional results on continuation randomness}\label{app:noise}

\begin{figure}[t]
\centering
\includegraphics[width=\linewidth]{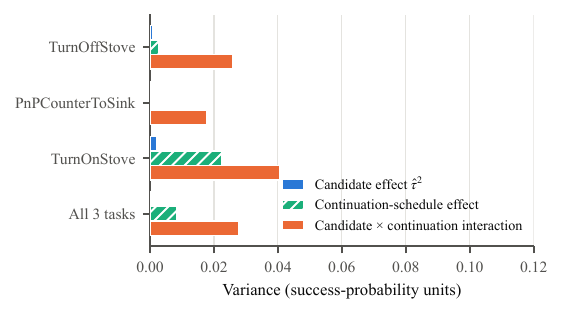}
\caption{\textbf{Variance decomposition of outcomes on E3}, the protocol of \citet{yuan2026beyond} (16 native candidates, continuation seeds shared across candidates): the candidate effect is negligible next to the candidate$\times$continuation interaction. The headroom versus the number of candidates on the same data is shown in \cref{fig:noise}.}
\label{fig:noise_full}
\end{figure}

\paragraph{RoboCasa.} \Cref{fig:noise_full} shows the variance decomposition per task. The pattern also holds on the task on which \citet{yuan2026beyond} found most of their mixed decision points: on TurnOffStove, $\OL=-0.003$. In E3, 19\% of decision points have mixed outcomes under the first schedule. A sensitivity analysis declared before any outcome was examined, which removes two trunks flagged by our rendering checks, leaves the ratio and the permutation test unchanged.

\paragraph{LIBERO-Plus.} \Cref{tab:liberoplus} summarizes the native and structured proposals of the four WAMs. On native candidates and under the protocol of \citet{yuan2026beyond} (first continuation only, all decision points), the single-continuation oracle reproduces a gain of 3--5 points over uniform selection, and 9--12\% of decision points have mixed outcomes; with the four second-stage continuations, $\Oone$ is 3.1--7.3 points (\cref{tab:liberoplus}). The candidate effect is at most 0.003, against an interaction of 0.015--0.026; for Streaming-WAM, it is indistinguishable from zero ($\tauhat=-0.0003$), and its held-out oracle lies at the center of the permutation null ($p=0.81$).

\begin{table}[t]
\centering\footnotesize
\setlength{\tabcolsep}{1.5pt}
\caption{LIBERO-Plus summary. Native: 16 native candidates with shared continuation schedules (E5). Structured: policy chunk $+$ 6 axis offsets at $\kappa=0.5$ (E6; E11 for 192 new states). Held-out share: $\OL/\Oone$ with three outcomes per candidate, whose expectation is a lower bound on the selectable share. All test quantities except the four-model pool were pre-registered; 95\% CIs from a bootstrap clustered by state.}
\label{tab:liberoplus}
\resizebox{\linewidth}{!}{\begin{tabular}{@{}lccccc@{}}
\toprule
& \multicolumn{2}{c}{Native} & \multicolumn{2}{c}{Structured, $\kappa=0.5$} \\
\cmidrule(lr){2-3}\cmidrule(l){4-5}
Model & $\Oone$ & Held-out share & Oracle $-$ chunk & Share \\
\midrule
Cosmos Policy & 0.073 & 0.15 $[0.02,0.30]$ & $+0.112\,[0.046,0.196]$ & 0.95 \\
Flex-$\pi$ & 0.047 & 0.18 $[-0.14,0.48]$ & $+0.045\,[0.006,0.096]$ & 0.79 \\
Fast-WAM & 0.073 & 0.09 $[-0.05,0.33]$ & $+0.088\,[0.034,0.157]$ & 0.86 \\
Streaming-WAM & 0.031 & $-0.09$ $[-0.23,0.05]$ & $+0.045\,[-0.010,0.116]$ & 0.78 \\
Streaming-WAM, E11 & & & $+0.052\,[0.026,0.083]$ & 0.80 \\
\midrule
Pooled, first three & 0.064 & 0.13 $[0.02,0.26]$ & & \\
Pooled, all four & 0.054 & 0.10 $[-0.01,0.21]$ & & \\
\bottomrule
\end{tabular}}
\end{table}

\paragraph{Judge-free selection.} KeyStone~\citep{dai2026keystone} returns the medoid of the largest cluster of sampled chunks. Because every native candidate in our data has repeated continuations, its choice can be scored without bias. \Cref{tab:keystone} reports KeyStone with its default settings on all native-candidate sets; the analysis plan was fixed before the chunks were exported. In the main analysis, no estimate differs from zero. Of the twelve reference and sensitivity intervals in the table, two exclude zero (the global medoid on E3 and KeyStone on executed actions for Flex-$\pi$); computed on executed actions, the global medoid also exceeds zero on E3 ($+0.027$) and for Fast-WAM ($+0.019$). Given the number of comparisons, we treat these results as exploratory; they are small and positive, as expected if consensus selection mainly avoids outlying samples. At a single decision point, the consensus candidate is therefore no better than a random one, consistent with native candidates differing mostly by continuation noise, and our per-decision estimates bound its effect at about 0.4 points per query on LIBERO-Plus. The unimodality guard fires at 10 of the 172 decision points (first initialization).

\begin{table*}[t]
\centering\footnotesize
\setlength{\tabcolsep}{4pt}
\caption{Judge-free selection on native candidates. KeyStone~\citep{dai2026keystone} with its default settings (L2 distance, unimodality threshold 0.3, two clusters, and the medoid of the larger cluster; mean over 20 seeded k-means initializations), applied offline to the 16 native candidates of each second-stage decision point. Gain: success rate of the chosen candidate minus the candidate mean, with two-stage weights; 95\% CIs from a bootstrap clustered by state. The main analysis uses the action chunk of the model (for Flex-$\pi$, the seven dimensions that reach the environment); the executed actions serve as a sensitivity analysis.}
\label{tab:keystone}
\begin{tabular}{@{}lccccc@{}}
\toprule
& & \multicolumn{2}{c}{KeyStone} & & \\
\cmidrule(lr){3-4}
Set & DPs & Model chunk & Executed actions & Global medoid & Held-out oracle \\
\midrule
E3 (RoboCasa, Cosmos Policy) & 36 & $+0.006\,[-0.002,0.015]$ & $+0.010\,[-0.005,0.029]$ & $+0.024\,[0.002,0.050]$ & $+0.009\,[-0.003,0.026]$ \\
E5, Cosmos Policy & 46 & $-0.007\,[-0.021,0.005]$ & $-0.006\,[-0.018,0.005]$ & $-0.010\,[-0.025,0.003]$ & $+0.011\,[0.002,0.024]$ \\
E5, Flex-$\pi$ & 46 & $-0.002\,[-0.011,0.006]$ & $+0.007\,[0.002,0.013]$ & $+0.010\,[-0.002,0.025]$ & $+0.008\,[-0.006,0.027]$ \\
E5, Fast-WAM & 44 & $-0.003\,[-0.021,0.013]$ & $+0.014\,[-0.000,0.029]$ & $-0.007\,[-0.031,0.014]$ & $+0.007\,[-0.004,0.020]$ \\
E5, pooled & 136 & $-0.004\,[-0.012,0.004]$ & $+0.005\,[-0.002,0.011]$ & $-0.002\,[-0.013,0.008]$ & $+0.009\,[0.002,0.017]$ \\
E5$^{+}$ (17 continuations) & 46 & $-0.003\,[-0.011,0.004]$ & $-0.003\,[-0.011,0.004]$ & $-0.004\,[-0.013,0.003]$ & $+0.019\,[0.005,0.040]$ \\
\bottomrule
\end{tabular}
\end{table*}

\section{Additional results on structured proposals}\label{app:proposals}

\begin{figure*}[t]
\centering
\includegraphics[width=\linewidth]{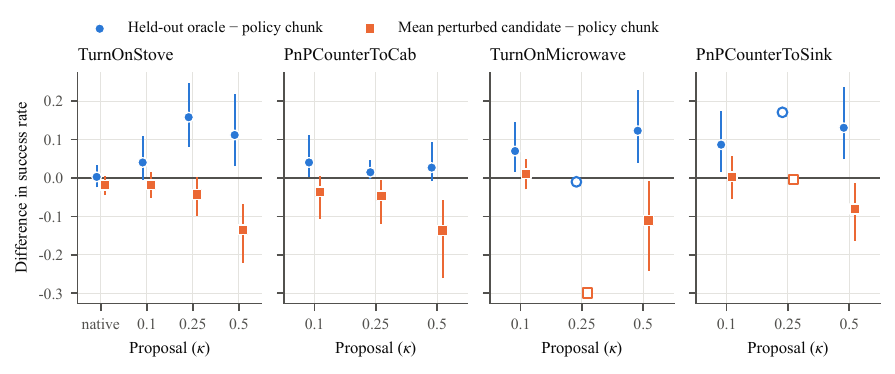}
\caption{\textbf{Headroom over the policy chunk on RoboCasa.} Blue: held-out oracle minus the policy chunk. Orange: the average perturbed candidate minus the policy chunk. Structured perturbations create selectable headroom on TurnOnStove, TurnOnMicrowave, and PnPCounterToSink; because the average perturbed candidate is worse than the policy chunk, the gain is available only to a selector. Hollow markers: 5 states, without error bars.}
\label{fig:headroom}
\end{figure*}

\Cref{fig:headroom} shows the headroom over the policy chunk for every RoboCasa task and scale. Increasing the sampling variance of the policy (the \texttt{use\_variance\_scale} option of Cosmos Policy, which randomly enlarges the noise levels of its sampler) does not create headroom for native candidates on TurnOnStove, which identifies the proposal, not the sampling temperature, as the source of headroom. The size of the headroom depends on the task: it lies between $+0.112$ and $+0.158$ on TurnOnStove, TurnOnMicrowave, and PnPCounterToSink and is $+0.027$ on PnPCounterToCab. On TurnOnStove, the direction main effect accounts for 2--6\% of $\tauhat$, and a fixed direction chosen on other states does not improve on the policy chunk. On LIBERO-Plus, the average perturbed candidate at $\kappa=0.5$ is 9.5, 11.5, 7.0, and 8.7 points below the policy chunk for Cosmos Policy, Fast-WAM, Flex-$\pi$, and Streaming-WAM, respectively.

\paragraph{Where the headroom lies.} \Cref{tab:wherefail} splits the held-out oracle at $\kappa=0.5$ by whether the policy chunk fails in at least one of the three continuations used for selection; the split never uses the scoring continuation. In all four WAMs, the gain over the policy chunk comes from these cases, and elsewhere it is close to zero. Streaming-WAM has the strongest policy chunk at its decision points and the fewest such cases, so its average gain is the smallest, although its gain where the chunk fails is as large as that of the other models; with three times as many decision points, E11 resolves this average gain. Its offsets move the end effector by 7.0--9.1\,cm by the end of the executed chunk; clipping affects 31\% of executed steps under $+x$, 26\% under $+y$, and at most 14\% under the other directions.

\begin{table}[t]
\centering\footnotesize
\setlength{\tabcolsep}{2.5pt}
\caption{\textbf{Where the structured headroom lies} (E6 and E11, $\kappa=0.5$; exploratory). Policy: success of the policy chunk. Fails: share of pairs of a decision point and a scoring continuation in which the policy chunk fails in at least one of the other three continuations. Gain: held-out oracle minus the policy chunk on the scoring continuation, split by this condition; 95\% CIs from a bootstrap clustered by state.}
\label{tab:wherefail}
\begin{tabular}{@{}lcccc@{}}
\toprule
Model & Policy & Fails & Gain where it fails & Elsewhere \\
\midrule
Cosmos Policy & 0.75 & 26\% & $+0.44\,[0.20,0.88]$ & $-0.004$ \\
Flex-$\pi$ & 0.75 & 25\% & $+0.21\,[0.07,0.48]$ & $-0.010$ \\
Fast-WAM & 0.66 & 41\% & $+0.21\,[0.08,0.39]$ & $+0.006$ \\
Streaming-WAM & 0.85 & 17\% & $+0.38\,[0.13,0.83]$ & $-0.022$ \\
Streaming-WAM, E11 & 0.85 & 19\% & $+0.27\,[0.14,0.43]$ & $+0.003$ \\
\bottomrule
\end{tabular}
\end{table}

\section{Additional results on the imagination}\label{app:imagination}

\paragraph{Value heads.} \Cref{tab:value} reports the state-level accuracy and the within-decision ranking of the public and fine-tuned value heads. All CIs of the within-decision Spearman correlation and of the selection gain include zero, except that of the 10K value head on perturbed candidates ($\rho=+0.18\,[0.004,0.356]$). Common random numbers make the ranking of the value head stable across seeds; deployed BoN instead draws independent seeds per candidate, and in that setting, the seed-to-seed noise of the value exceeds its differences between candidates by an order of magnitude, and the argmax agrees between two independent scorings only 15--27\% of the time (chance: 1/6).

\begin{table}[t]
\centering\footnotesize
\setlength{\tabcolsep}{2.5pt}
\caption{\textbf{Value heads value states, not candidates} (E1, common random numbers, $s'=s_{t_b+16}$). AUROC: state value versus final success. $\rho$: within-decision Spearman correlation between $\bar V$ and $\phat$. Gain: success gain of selecting by $\bar V$ over the candidate mean. Cons.: within-decision rank agreement across three value seeds.}
\label{tab:value}
\begin{tabular}{@{}lccccccc@{}}
\toprule
& & \multicolumn{3}{c}{Native} & \multicolumn{3}{c}{Perturbed ($\kappa{=}0.1$)} \\
\cmidrule(lr){3-5}\cmidrule(l){6-8}
Value head & AUROC & $\rho$ & Gain & Cons. & $\rho$ & Gain & Cons. \\
\midrule
Public & 0.53 & $+0.05$ & $-0.001$ & 0.72 & $+0.02$ & $-0.067$ & 0.94 \\
Tuned 5K & 0.87 & $+0.03$ & $-0.001$ & 0.74 & $+0.02$ & $-0.006$ & 0.78 \\
Tuned 10K & 0.90 & $+0.03$ & $+0.008$ & 0.84 & $+0.18$ & $+0.001$ & 0.86 \\
Tuned 15K & 0.90 & $+0.02$ & $+0.003$ & 0.72 & $+0.11$ & $+0.001$ & 0.84 \\
\bottomrule
\end{tabular}
\end{table}

\begin{figure*}[t]
\centering
\includegraphics[width=\linewidth]{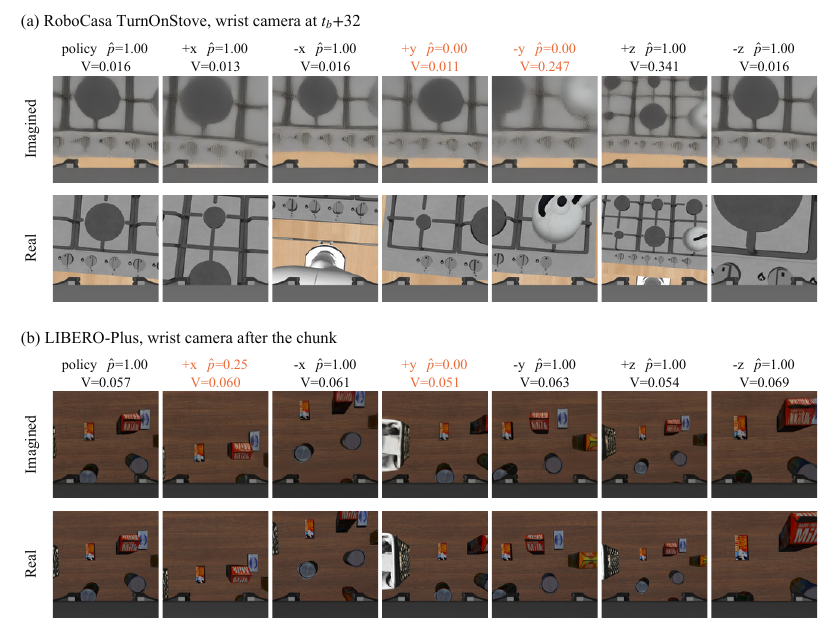}
\caption{\textbf{Imagined versus real futures of seven sibling candidates} (Cosmos Policy, $\kappa=0.5$; for each benchmark, the first decision point by index among those with the largest outcome range). Columns: the policy chunk and its six axis offsets; $\hat p$: success rate over four continuations (orange: $\hat p<0.5$); $V$: value head on the imagined future. On RoboCasa, the imagined frames reproduce the direction-specific consequences, such as the kettle entering the view under $-y$ and the higher viewpoint under $+z$, at lower sharpness and with smaller displacements than the real ones. On LIBERO-Plus, the imagined views closely match the real ones, including the basket edge that appears only under the failing $+y$ offset.}
\label{fig:qual}
\end{figure*}

\paragraph{Fidelity of the imagination.} The imagined futures are faithful to the action itself: the commanded 2\,cm displacement direction is decodable from them with $R^2=0.43$--$0.45$, at least as well as from real futures. \Cref{fig:qual} shows imagined and real futures side by side for one decision point per benchmark, chosen by a fixed rule; the imagination reproduces the consequences that distinguish sibling candidates, including those of the candidates that fail. The probe requires a model that conditions its predicted future on a supplied action; the released checkpoint of Flex-$\pi$ generates its future jointly with its own action, and Fast-WAM does not generate the future at test time, so the probe is applied to Cosmos Policy and Streaming-WAM.

\paragraph{When the information appears.} Readability depends on the task and on the scale of the perturbation, and the protocol shows why. On TurnOnStove, 2\,cm perturbations produce reliable outcome differences (split-half Spearman correlation across continuations of $+0.39\,[0.16,0.64]$) that neither the real nor the imagined future at $t_b+16$ or $t_b+32$ reveals, which indicates that the outcome is decided later in the closed loop. At 5--10\,cm, the consequence is visible in both. On TurnOnMicrowave, the consequence is visible already at 2\,cm ($\rho=+0.31$). In every condition in which a probe of the real near future ranks candidates, so does the probe of the imagined one.

\paragraph{Streaming-WAM.} The pre-registered probe reads the final-block tokens of the latent frame that covers steps 4--16, averaged over three imagination seeds (\cref{app:protocol}); both arms meet the premise (30 and 41 states). The probe of the imagined future reaches $\rho=0.079\,[-0.108,0.286]$ ($p=0.32$) at $\kappa=0.1$ and $0.110\,[-0.014,0.243]$ ($p=0.10$) at $\kappa=0.5$, and the same probe of the real future reaches $-0.059$ ($p=0.65$) and $0.162\,[-0.006,0.322]$ ($p=0.037$); the paired difference, pooled over both arms, is $-0.005\,[-0.146,0.129]$, within the pre-registered equivalence band. On the 192 new states of E11 (195 decision points, 146 states), the pre-registered repetition at $\kappa=0.5$ passes for both features: the final-block tokens reach $\rho=0.157\,[0.077,0.232]$ ($p=0.0015$) for the imagined and $0.190\,[0.113,0.263]$ ($p=0.0005$) for the real future, and the wrist-camera spatial mean reaches $0.102\,[0.029,0.173]$ ($p=0.017$) and $0.195\,[0.120,0.268]$ ($p=0.0005$). The paired difference of the token probes is $-0.033\,[-0.091,0.024]$, within the pre-registered band, whereas that of the wrist-camera summary is $-0.094\,[-0.178,-0.009]$ (descriptive), so the full token set carries the imagined consequence more completely than its spatial mean. In exploratory readouts of lower-dimensional features at $\kappa=0.5$, the spatial mean of the wrist-camera tokens ranks candidates with $\rho=0.207\,[0.089,0.323]$ ($p=0.011$) for the imagined and $0.247\,[0.091,0.382]$ ($p=0.001$) for the real future, and that of the third-person camera with $0.138$ ($p=0.062$) and $0.207$ ($p=0.010$); the action-expert tokens of the supplied prefix reach $0.125$ ($p=0.069$).

\paragraph{Future versus action frames.} \Cref{tab:ablation} reports, for every perturbation condition, the readout of the imagination pass split by latent frame (final transformer block, mean over three imagination seeds; frames 5, 6--9, and 10 on RoboCasa and frames 4, 5--7, and 8 on LIBERO for the action, the future, and the value, respectively). ``Fut $|$ act'' residualizes the future-frame prediction within each decision point on the action-frame prediction, and ``act $|$ fut'' is the reverse; ``fut $|$ state$\times$dir'' residualizes on a probe of the current-state representation (16 principal components of the current-frame tokens of the policy chunk) interacted with the commanded direction. All predictions are first residualized on the leave-one-state-out global direction. Bold: permutation $p<0.05$ (500 refits). Pooled values in the main text use a bootstrap clustered by state. Pooled over the twelve conditions, the future frames retain $0.13\,[0.08,0.18]$ beyond the state$\times$direction baseline.

\begin{table*}[t]
\centering\small
\caption{Readout by latent frame of the imagination pass (within-decision Spearman correlation with the outcome).}
\label{tab:ablation}
\begin{tabular}{@{}lccccccc@{}}
\toprule
Condition & DPs & Future & Action & Fut $|$ act & Act $|$ fut & Fut $|$ state$\times$dir & Value frame \\
\midrule
TurnOnStove $\kappa{=}0.1$ & 40 & $-$0.19 & $-$0.09 & $-$0.15 & +0.17 & $-$0.25 & $-$0.01 \\
TurnOnStove $\kappa{=}0.25$ & 53 & \textbf{+0.31} & \textbf{+0.17} & +0.08 & +0.05 & +0.14 & \textbf{+0.22} \\
TurnOnStove $\kappa{=}0.5$ & 59 & \textbf{+0.37} & \textbf{+0.19} & \textbf{+0.16} & $-$0.05 & \textbf{+0.29} & \textbf{+0.19} \\
PnPCounterToCab $\kappa{=}0.1$ & 19 & +0.20 & +0.07 & +0.14 & $-$0.06 & +0.06 & \textbf{+0.30} \\
PnPCounterToCab $\kappa{=}0.25$ & 20 & +0.23 & +0.03 & +0.25 & $-$0.15 & +0.16 & +0.04 \\
PnPCounterToCab $\kappa{=}0.5$ & 23 & +0.06 & +0.11 & $-$0.14 & +0.18 & +0.03 & $-$0.07 \\
TurnOnMicrowave $\kappa{=}0.1$ & 26 & \textbf{+0.31} & +0.16 & +0.22 & $-$0.09 & +0.25 & +0.09 \\
TurnOnMicrowave $\kappa{=}0.5$ & 39 & \textbf{+0.54} & \textbf{+0.43} & \textbf{+0.19} & $-$0.06 & \textbf{+0.19} & \textbf{+0.39} \\
PnPCounterToSink $\kappa{=}0.1$ & 32 & +0.17 & $-$0.00 & +0.17 & $-$0.13 & +0.05 & +0.12 \\
PnPCounterToSink $\kappa{=}0.5$ & 45 & \textbf{+0.17} & +0.07 & \textbf{+0.19} & $-$0.00 & +0.13 & +0.10 \\
LIBERO-Plus $\kappa{=}0.1$ & 38 & \textbf{+0.26} & +0.19 & +0.16 & +0.06 & +0.16 & +0.01 \\
LIBERO-Plus $\kappa{=}0.5$ & 69 & \textbf{+0.20} & \textbf{+0.24} & +0.06 & \textbf{+0.13} & \textbf{+0.17} & $-$0.09 \\
\midrule
Pooled & 360 & 0.23 & 0.15 & 0.11 & 0.02 & 0.13 & 0.11 \\
\bottomrule
\end{tabular}
\end{table*}

\section{Selectors}\label{app:selectors}

\paragraph{Training data.} E7 contains 320 LIBERO-Plus states (Objects Layout and Robot Initial States), which are disjoint from all other sets and from the reserved closed-loop evaluation list, with decision points at $t_b\in\{0,32,64\}$. Each decision point has 13 candidates (the policy chunk and six axis offsets at each of $\kappa=0.1$ and $0.5$), one first-stage continuation per candidate, and two further independent continuations for the 696 decision points (304 states) that enter the second stage.

\paragraph{Ranking with abstention.} This selector shows why the decision to deviate needs its own model. Its features are the final-block tokens of the imagined future frames of Cosmos Policy (one imagination seed), centered within each decision point and standardized, together with a centered one-hot code of the candidate. A weighted dual ridge model predicts $\phat$, and the selector deviates from the policy chunk only when the best predicted candidate exceeds the policy chunk by more than $\tau$; the ridge strength $a$ and $\tau$ were chosen by five-fold cross-validation over states on E7, and the final model was evaluated once on E6. Because an average offset is worse than the policy chunk, every finite threshold has a negative cross-validated gain, and cross-validation selects $\tau=\infty$ (\cref{tab:p10}); the same holds when training on the first 80, 160, or 240 states of each cell, and a model with only the one-hot code (a fixed direction) never deviates either, because the policy chunk has the highest average outcome. The gain improves monotonically with $a$ up to the edge of the pre-registered grid ($a\le10^2$), which is narrower than the ranker grid of the fail-gated selector ($a=10^2$--$10^6$); in an exploratory extension to $a=10^3$--$10^6$, some finite thresholds reach a cross-validated gain of up to $+0.02$ (a maximum over the same folds, without nested cross-validation or a CI), so the two selectors were not tuned with matched grids. Within the pre-registered grid, ranking alone never deviates from a strong policy chunk, and a separate failure gate supplies the decision to deviate. Because this selector never deviates, its closed-loop evaluation was not run.

\begin{table}[t]
\centering\footnotesize
\setlength{\tabcolsep}{2.5pt}
\caption{Ranking with abstention: cross-validated gain over the policy chunk on E7 (696 decision points, 13 candidates) over the pre-registered grid of $a$ and $\tau$.}
\label{tab:p10}
\begin{tabular}{@{}lccccccc@{}}
\toprule
$a$ & $\tau=0$ & $0.02$ & $0.05$ & $0.1$ & $0.15$ & $0.2$ & $\infty$ \\
\midrule
0.1 & $-0.055$ & $-0.055$ & $-0.055$ & $-0.054$ & $-0.051$ & $-0.047$ & 0 \\
1 & $-0.053$ & $-0.052$ & $-0.054$ & $-0.053$ & $-0.050$ & $-0.047$ & 0 \\
10 & $-0.043$ & $-0.043$ & $-0.042$ & $-0.042$ & $-0.037$ & $-0.035$ & 0 \\
100 & $-0.030$ & $-0.030$ & $-0.028$ & $-0.024$ & $-0.021$ & $-0.019$ & 0 \\
\bottomrule
\end{tabular}
\end{table}

\paragraph{Fail-gated selector.} The gate is an L2-regularized logistic regression that predicts the failure of the policy chunk from the imagination of the WAM for that chunk; it is trained on the first-stage outcome of all 944 E7 decision points and on the two second-stage outcomes where they are available (with two-stage weights). The ranker is the dual ridge model described above, centered within the candidate set. With the gate score $q(a)$ of the policy chunk $a$, the threshold $\theta$, the offset set $\mathcal C$, and the ranker score $r(c)$, the selector executes $\arg\max_{c\in\mathcal C}r(c)$ if $q(a)>\theta$ and $a$ otherwise. The pre-declared grid covered the gate features (all nine latent frames of the imagination pass, or the current-observation frames together with the value), the gate regularization ($C\in\{10^{-4},\dots,1\}$), the ranker strength ($a\in\{10^2,\dots,10^6\}$), the ranker training set (all second-stage decision points, or only those at which the chunk failed), the candidate set (6 or 12 offsets), and the threshold (50th--90th percentile of the gate score), for 1{,}000 configurations in total. Cross-validation over the same five state folds as for ranking with abstention selected all nine frames, $C=10^{-4}$, all decision points, 12 offsets, $a=10^5$, and the 60th percentile (cross-validated gain $+0.028$); the same gate with a random offset gives $-0.021$, and with the best fixed direction at most $+0.011$. Nested cross-validation (5 outer $\times$ 5 inner folds) estimates the gain of the whole selection procedure at $+0.025\,[0.009,0.041]$, with deviation at 35\% of held-out decision points and a gate AUROC of 0.88; by category, the gain is $+0.055$ on robot initial states and $-0.005$ on object layouts. On E6, inspected once after freezing and only descriptively, the selector gains $+0.020\,[-0.028,0.076]$ in the $\kappa=0.5$ pool. On E9, the frozen model scores the imagination of the policy chunk at every decision point; only where the gate fires are the twelve offsets imagined with the same noise seed and ranked. In closed loop, the selector call restores the global random state after imagination, so that the reseeding performed by the image encoder does not alter the sensor noise of the environment; an equivalence check on ten E7 decision points confirmed that the online call reproduces the offline features, gate score, and choice bit for bit.

\paragraph{Further results on E9 and E8.} The gate trades rescues against losses: 42\% of deviations on E9 occur where the policy chunk always fails, and where it always succeeds, deviating costs 7.6 points. The gain is largest at the first decision point of an episode ($+0.036$) and concentrates on robot initial-state perturbations ($+0.040$), as in development. The choice is state-dependent: the upward offset at $\kappa=0.5$ is the most frequent choice (29\% of deviations), yet on the 39 deviations whose seeded random offset happened to be the upward one, which form a random subsample, the upward offset gains $+0.006$ per deviation, against $+0.115$ for the selector on the same decision points (exploratory). On E8, 127 pairs succeed only with the selector and 59 only with the policy; by perturbation category, success rises from 65.1\% to 75.8\% under robot initial states and is 87.4\% against 88.6\% for the policy ($-1.3\,[-2.8,0.1]$ points) under object layouts, where the policy already succeeds most of the time. E9 and E8 contain new episodes of the training tasks: their initial observations differ from every training episode, while most robot initial-state episodes share their object layout with a training episode (194 of 200 in E9, 235 of 240 in E8). Transfer to tasks absent from training is therefore tested separately, on E10.

\paragraph{What the selector relies on.} We repeated the nested cross-validation of the fail-gated selector on E7 with the same folds and pipeline, changing one ingredient at a time (\cref{tab:selablation}); our re-implementation reproduces the frozen model to machine precision and the nested estimate of $+0.025$ exactly. The gate reads a state-level judgment from the representation of the current scene: current-observation frames alone match the full representation, whereas the scalar output of the value head recovers $+0.007$, so the representation holds failure information that the value projection discards. The ranker draws its gain from state-dependent choices, since a fixed direction learned on the same data gains nothing. Within tasks, a ranker that sees only the current state and the commanded direction (64 principal components) comes close to the full representation ($+0.019$ against $+0.025$). When entire tasks are held out (outer and inner folds grouped by the 40 base tasks, as for E10), the full model gains $+0.005\,[-0.008,0.018]$, the ranker that reads the imagined future frames $+0.004\,[-0.011,0.018]$, and the state-by-direction ranker $+0.007\,[-0.002,0.016]$, so offline cross-validation over 40 tasks cannot separate the rankers; rankers that read the imagination were selected in all four folds of the unseen-task test. Transfer to unseen tasks was therefore tested directly in closed loop on E10.

\begin{table*}[t]
\centering\small
\caption{What the fail-gated selector relies on. Nested cross-validation on E7 (696 decision points; the same outer folds as for the frozen selector), changing one ingredient at a time; gain over the policy chunk with two-stage weights; 95\% CIs from a bootstrap clustered by state (by base task for task folds). Diff.: full model minus the variant, paired over decision points.}
\label{tab:selablation}
\begin{tabular}{@{}lcc@{}}
\toprule
Variant & Gain & Diff. \\
\midrule
Full model (recipe of the frozen selector) & $+0.025\,[0.009,0.041]$ & --- \\
\midrule
Gate: current-observation frames only & $+0.025\,[0.009,0.041]$ & $+0.000\,[-0.002,0.002]$ \\
Gate: imagined future frames only & $+0.025\,[0.009,0.042]$ & $+0.000\,[-0.008,0.007]$ \\
Gate: value-head scalar only & $+0.007\,[-0.004,0.019]$ & $+0.018\,[0.004,0.031]$ \\
\midrule
Ranker: imagined future frames only & $+0.020\,[0.005,0.035]$ & $+0.005\,[-0.004,0.014]$ \\
Ranker: action frame only & $+0.018\,[0.004,0.033]$ & $+0.007\,[-0.006,0.019]$ \\
Ranker: all but the action frame & $+0.026\,[0.011,0.043]$ & $-0.002\,[-0.007,0.003]$ \\
Ranker: state $\times$ direction, 64 PCs & $+0.019\,[0.005,0.033]$ & $+0.006\,[-0.008,0.020]$ \\
Ranker: state $\times$ direction, 16 PCs & $+0.011\,[-0.001,0.024]$ & $+0.014\,[0.000,0.028]$ \\
Ranker: fixed direction (one-hot only) & $-0.002\,[-0.008,0.004]$ & $+0.026\,[0.012,0.042]$ \\
Ranker: random offset & $+0.003\,[-0.001,0.007]$ & $+0.022\,[0.007,0.037]$ \\
\midrule
Task folds: full model & $+0.005\,[-0.008,0.018]$ & --- \\
Task folds: imagined future frames (ranker) & $+0.004\,[-0.011,0.018]$ & \\
Task folds: state $\times$ direction, 16 PCs & $+0.007\,[-0.002,0.016]$ & \\
Task folds: fixed direction & $+0.000\,[-0.005,0.004]$ & \\
\bottomrule
\end{tabular}
\end{table*}

\begin{figure}[t]
\centering
\includegraphics[width=\linewidth]{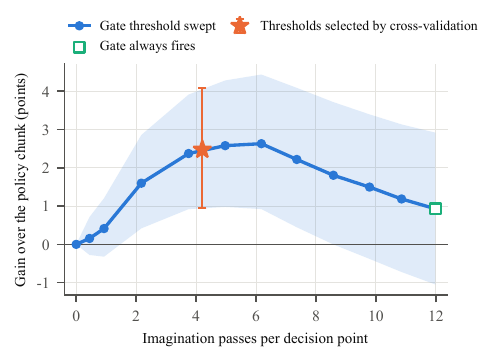}
\caption{\textbf{Gain versus imagination cost} (E7, nested cross-validation; exploratory). The gate and the ranker of each outer fold are those selected by the inner cross-validation, and only the gate threshold is swept. Cost: imagination passes per decision point, that is, twelve times the share of decision points at which the gate fires. Star: the thresholds selected by cross-validation; square: the gate always fires. Band and bar: 95\% CIs from a bootstrap clustered by state.}
\label{fig:cost}
\end{figure}

\paragraph{Compute.} \Cref{fig:cost} sweeps the gate threshold in the nested cross-validation on E7, keeping the gate and the ranker selected in each outer fold. Imagining the twelve offsets at every decision point gains $+0.009\,[-0.010,0.029]$. As the gate becomes more selective, the gain rises to its maximum of $+0.026\,[0.009,0.044]$ when the gate fires at about half of the decision points, and the thresholds selected by cross-validation reach $+0.025\,[0.009,0.041]$ while the gate fires at 35\% of them. Firing at 18\% of the decision points, that is, 2.2 imagination passes per decision point, still retains $+0.016\,[0.004,0.029]$. The gate thus raises the gain and lowers the cost at the same time: it spends imagination where deviating helps and saves it where every deviation can only lose. Each imagination pass is one run of the policy network with the candidate chunk injected; in closed loop, the frozen selector added 17.5 passes per episode to 12 policy queries (\cref{sec:selector}).

\begin{figure*}[t]
\centering
\includegraphics[width=\linewidth]{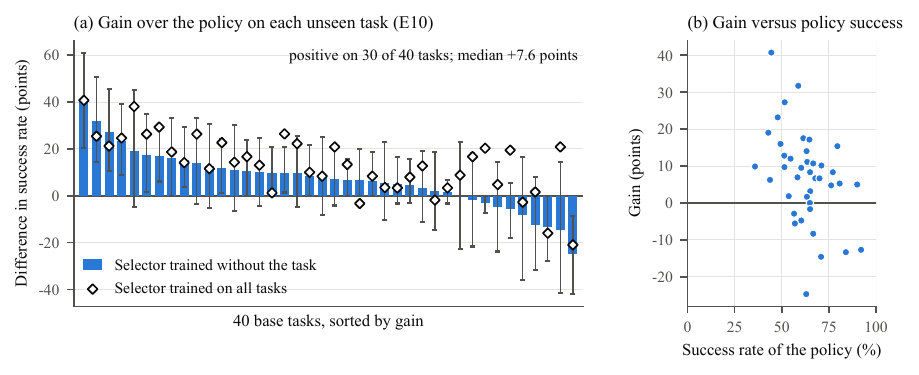}
\caption{\textbf{Closed-loop gain on each unseen task} (E10: Cosmos Policy on LIBERO-Plus, robot initial states; 10--33 episodes per task, three replicates per episode with common random numbers). (a)~Success rate of the selector that never saw the task minus that of the policy (bars; 95\% CIs from a bootstrap over the episodes of the task), and the same difference for the selector trained on all tasks (diamonds); tasks are sorted by the first difference. (b)~The first difference against the success rate of the policy on the same task.}
\label{fig:e10tasks}
\end{figure*}
\begin{figure*}[t]
\centering
\includegraphics[width=\linewidth]{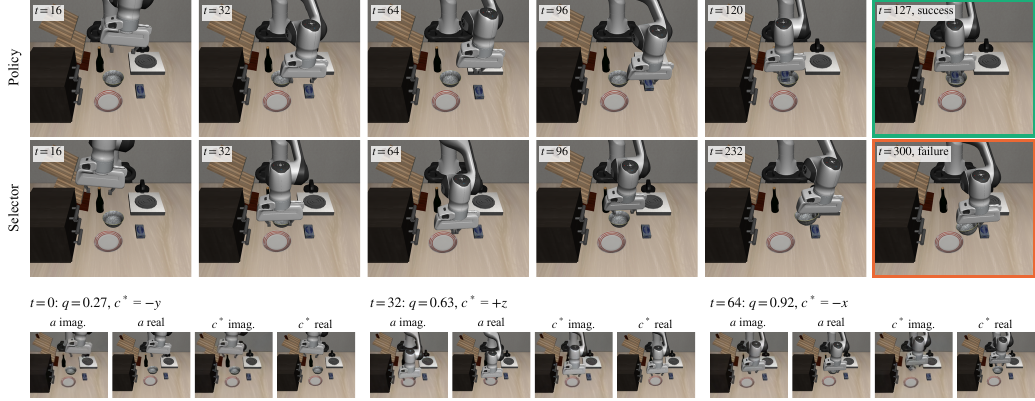}
\caption{\textbf{Closed-loop counterexample on an unseen task} (E10: ``put the cream cheese in the bowl''; the first, in hash order, of the 26 episodes in which the policy succeeds and the selector fails in all three replicates; both runs share all seeds). Top: the policy succeeds; the selector deviates at $t=0$, 32, and 64 and times out. Bottom: at each deviation, imagined frames and the real frames 16 steps after executing the policy chunk $a$ or the selected offset $c^{*}$ from the state of the selector run. Because the two runs separate at $t=0$, the real frames for $a$ at $t=32$ and 64 come from a 16-step branch from that state, not from the policy run.}
\label{fig:closedloop_harm}
\end{figure*}

\paragraph{Selectors for unseen tasks.} For each of the four task folds, the recipe of the fail-gated selector was re-run on E7 without the tasks of that fold (both perturbation categories removed; 230--248 states), with cross-validation folds grouped by base task and the ranker features as an additional grid dimension (the full representation, the imagined future frames, or the current-state-by-direction baseline). The selected configurations use the full-representation gate in all folds; the ranker reads the imagined future frames in folds 0 and 1 and the full representation in folds 2 and 3. The best cross-validated gains of the full-representation, imagined-future, and state-by-direction rankers are $+0.016/+0.021/+0.013$, $+0.019/+0.019/+0.006$, $+0.018/+0.012/+0.009$, and $+0.007/+0.002/+0.001$ in folds 0--3. Our re-implementation reproduces all four frozen models from E7 without the tasks of each fold to within $10^{-7}$ (gate) and $10^{-15}$ (ranker). In E10, every episode in which a selector never deviates is identical to its policy counterpart (1{,}013 episodes verified). The closed-loop gain of the selector that never saw the task is $+4.9$, $+12.4$, $+10.0$, and $+1.8$ points in folds 0--3; the selector trained on all tasks exceeds it by $+6.2\,[4.1,8.4]$ points. \Cref{fig:e10tasks} shows the gain on each task: it is positive on 30 of the 40 tasks, with a median of $+7.6$ points, and it is larger on tasks at which the policy succeeds less often (Pearson $r=-0.44$ across tasks; exploratory). \Cref{fig:closedloop,fig:closedloop_harm} show the first episodes, in hash order, in which the selector rescues the policy in all three replicates (62 episodes) and in which it harms the policy in all three (26 episodes).

\begin{table*}[t]
\centering\small
\caption{Offline selection on RoboCasa in the seven-candidate pool (policy chunk $+$ 6 offsets); gain over executing the policy chunk. Learned selectors are evaluated leave-one-state-out. Probe: global direction $+$ linear residual on the representation.}
\label{tab:bon}
\begin{tabular}{@{}llcccccc@{}}
\toprule
Task & $\kappa$ & States & Held-out oracle & Fixed direction & Imagined probe & Real $s_{t_b+32}$ probe & Value head \\
\midrule
TurnOnStove & 0.25 & 35 & $+0.158$ & $-0.006$ & $+0.008$ & $-0.006$ & $+0.014$ \\
TurnOnStove & 0.5 & 36 & $+0.112$ & $\phantom{+}0.000$ & $-0.017$ & $-0.070$ & $-0.090$ \\
PnPCounterToCab & 0.5 & 15 & $+0.027$ & $\phantom{+}0.000$ & $-0.055$ & $-0.098$ & $-0.226$ \\
PnPCounterToSink & 0.5 & 24 & $+0.131$ & $-0.052$ & $-0.020$ & $+0.008$ & $-0.020$ \\
TurnOnMicrowave & 0.1 & 18 & $+0.070$ & $+0.069$ & $+0.069$ & $+0.036$ & $+0.008$ \\
TurnOnMicrowave & 0.5 & 23 & $+0.123$ & $+0.032$ & $+0.097$ & $+0.093$ & $-0.144$ \\
\bottomrule
\end{tabular}
\end{table*}

\paragraph{Offline selection on RoboCasa.} \Cref{tab:bon} evaluates learned selectors in the seven-candidate pool that contains the policy chunk on RoboCasa, with 15--36 training states per task. At this data scale, gains over the policy chunk appear on TurnOnMicrowave, where a fixed correction direction recovers the full headroom at $\kappa=0.1$ and the imagined-future probe recovers 80\% of it at $\kappa=0.5$ (permutation $p=0.088$; the gain does not persist when entire scenes are held out), and remain within noise on TurnOnStove and PnPCounterToSink. Learned selection of state-dependent directions therefore needs training data at the scale of E7, 320 states, which is the setting of \cref{sec:selector}.

\section{HeadroomBench}\label{app:bench}

\begin{table}[t]
\centering\footnotesize
\setlength{\tabcolsep}{2.5pt}
\caption{\textbf{Reference results of the selector track} (E6 test split, Cosmos Policy; primary pool: policy chunk $+$ 6 offsets at $\kappa=0.5$, among which every selector chooses; 69 decision points, 48 states). Gain over the policy chunk with 95\% CIs from a bootstrap clustered by state. Dev.: share of decision points at which the selector deviates from the policy chunk. $^\dagger$E6 informed the design of this selector; its held-out evaluation is on E9.}
\label{tab:bench}
\begin{tabular}{@{}lcc@{}}
\toprule
Selector & Gain & Dev. \\
\midrule
Policy chunk & 0 & 0\% \\
Random candidate & $-0.081\,[-0.136,-0.033]$ & 86\% \\
Value head on the imagined future & $-0.096\,[-0.175,-0.022]$ & -- \\
Ranking with abstention & $0\,[0,0]$ & 0\% \\
Failure gate $+$ random offset$^\dagger$ & $-0.040\,[-0.083,-0.004]$ & 42\% \\
Failure gate $+$ learned ranker$^\dagger$ & $+0.020\,[-0.028,0.076]$ & 42\% \\
Held-out oracle (reference) & $+0.112\,[0.047,0.198]$ & -- \\
\bottomrule
\end{tabular}
\end{table}

\paragraph{Headroom track.} Given a policy with seeded queries and a simulator with state snapshots, the track measures the single-continuation oracle of a proposal, its held-out share together with the continuation-count curve of \cref{fig:precision}, its headroom over the policy chunk, and the variance decomposition, all with state-clustered CIs. The two-stage design adds continuations only where first-stage outcomes differ, plus a random audit quarter of the remaining points (about 30\% of decision points in total on E5), so the cost stays close to that of the standard protocol. The package contains branching adapters for Cosmos Policy, Flex-$\pi$, Fast-WAM, and Streaming-WAM on LIBERO-Plus and for Cosmos Policy on RoboCasa, the estimators, the rendering-integrity checks of \cref{app:protocol}, the linear readout of \cref{sec:imagination}, the fail-gated selectors with their frozen parameters and the code that runs them online, and tests that reproduce the headline numbers of this paper from the released outcome tables and representations.

\paragraph{Selector track.} At each decision point, a selector scores candidate chunks with the imagination of the WAM and chooses one; the score is the success gain over executing the policy chunk. The training split is E7 (320 states, 944 decision points), where every decision point offers 13 candidates, namely the policy chunk and twelve axis offsets at $\kappa=0.1$ and 0.5, with outcomes and imagined-future representations. The test split is E6 (56 disjoint states, four continuations per candidate), whose offsets at $\kappa=0.1$ and 0.5 were screened as separate arms; its primary pool is therefore the policy chunk and the six offsets at $\kappa=0.5$, among which a selector trained on 13 candidates chooses, and the package also scores all 13 candidates at the decision points where both arms were measured. Two lists of further episodes (E8 and E10) serve closed-loop evaluation; each has been used once, for the closed-loop tests of \cref{sec:selector}. \Cref{tab:bench} gives the reference results. The held-out oracle exceeds the policy chunk by 11 points, whereas selection by the value head and random selection both lose 8--10 points. The split thus rewards selectors that deviate selectively: the ranking signal in the imagination is real ($\rho\approx0.2$), but an average offset costs 8--10 points, so ranking with abstention stays with the policy chunk, and the fail-gated selector of \cref{sec:selector} gains $+0.020$ (E6 informed its design; its held-out evaluation is the one on E9).